\documentclass{article} 
\usepackage{iclr2024_conference,times}

\usepackage{amsmath,amsfonts,bm}

\def\eqref#1{equation~\ref{#1}}

\def\1{\bm{1}}

\DeclareMathAlphabet{\mathsfit}{\encodingdefault}{\sfdefault}{m}{sl}
\SetMathAlphabet{\mathsfit}{bold}{\encodingdefault}{\sfdefault}{bx}{n}

\usepackage[pagebackref=true,breaklinks=true,colorlinks,citecolor=gray]{hyperref}
\usepackage{xcolor}
\definecolor{citecolor}{HTML}{0071BC}
\hypersetup{colorlinks,linkcolor={red},citecolor={citecolor}}  

\usepackage{url}
\usepackage{soul}
\usepackage{booktabs}
\usepackage{amsmath,caption}
\usepackage{colortbl}
\usepackage{arydshln}
\usepackage{multirow}
\usepackage{multicol}

\usepackage{xspace}
\usepackage{adjustbox}
\usepackage{subcaption}
\usepackage{longtable}
\usepackage{makecell}
\usepackage{tabularx}
\usepackage{pifont}
\usepackage{colortbl}
\usepackage{array}
\usepackage{verbatim}
\usepackage{svg}
\usepackage{amsmath}
\usepackage[ruled,linesnumbered]{algorithm2e}
\usepackage{lipsum} 
\usepackage{graphicx} 
\usepackage{wrapfig} 
\usepackage{floatflt}
\usepackage{tcolorbox}
\usepackage{relsize}

\newcommand\ours{\textsc{Tree-Planner}\xspace}

\newcommand{\observation}{\textit{O}}

\newcommand{\states}{\textit{S}}

\newcommand{\actions}{\textit{A}}

\newcommand{\progprompt}{\textsc{ProgPrompt}\xspace}
\newcommand{\zeroshot}{\textsc{Zero-shot Planner}\xspace}
\newcommand{\sbs}{\textsc{Iterative-Planner}\xspace}
\newcommand{\localreplan}{\textsc{Local Replan}\xspace}
\newcommand{\globalreplan}{\textsc{Global Replan}\xspace}
\newcommand{\exec}{\textsc{Exec.}\xspace}
\newcommand{\sr}{\textsc{SR}\xspace}
\newcommand{\gcr}{\textsc{GCR}\xspace}
\newcommand{\cost}{\textsc{\$Cost}\xspace}
\newcommand{\nocorrection}{\textsc{No.EC}\xspace}

\title{\ours: Efficient Close-loop Task Planning with Large Language Models}
\author{
        \textbf{Mengkang Hu} $^\spadesuit$ \ \ 
        \textbf{Yao Mu} $^{\spadesuit}$ \ \ 
        \textbf{Xinmiao Yu}$^{\heartsuit}$ \ \ 
        \textbf{Mingyu Ding}$^*$$^\spadesuit$ \ \ 
        \textbf{Shiguang Wu}$^\diamondsuit$ \\ 
        \textbf{Wenqi Shao}$^{\clubsuit}$\ \
        \textbf{Qiguang Chen}$^{\heartsuit}$\ \
        \textbf{Bin Wang}$^{\diamondsuit}$\ \
        \textbf{Yu Qiao}$^{\clubsuit}$\ \
        \quad \textbf{Ping Luo}$^*$ \thanks{
        Corresponding authors: Mingyu Ding and Ping Luo (\{dingmyu, pluo.lhi\}@gmail.com).
        $^\spadesuit$The University of Hong Kong. $^\heartsuit$Harbin Institute of Technology. $^\diamondsuit$Noah’s Ark Laboratory. $^\clubsuit$Shanghai AI Laboratory.
        }$^\spadesuit$
}

\iclrfinalcopy 
\begin{document}

\maketitle
\begin{abstract}
This paper studies close-loop task planning, which refers to the process of generating a sequence of skills (a plan) to accomplish a specific goal while adapting the plan based on real-time observations.
Recently, prompting Large Language Models (LLMs) to generate actions iteratively has become a prevalent paradigm due to its superior performance and user-friendliness.
However, this paradigm is plagued by two inefficiencies: high token consumption and redundant error correction, both of which hinder its scalability for large-scale testing and applications.
To address these issues, we propose \ours, which reframes task planning with LLMs into three distinct phases: 
plan sampling,  action tree construction, and grounded deciding.
\ours starts by using an LLM to sample a set of potential plans before execution, followed by the aggregation of them to form an action tree.
Finally, the LLM performs a top-down decision-making process on the tree, taking into account real-time environmental information.
Experiments show that \ours achieves state-of-the-art performance while maintaining high efficiency.
By decomposing LLM queries into a single plan-sampling call and multiple grounded-deciding calls,
a considerable part
of the prompt are less likely to be repeatedly consumed. 
As a result, token consumption is reduced by 92.2\% compared to the previously best-performing model.
Additionally, by enabling backtracking on the action tree as needed, the correction process becomes more flexible, leading to a 40.5\% decrease in error corrections.
\end{abstract}

\section{Introduction}
\label{sec:intro}

\begin{wrapfigure}{r}{0.28\columnwidth}
\vspace{-0.6cm}
    \begin{center}
    \includegraphics[width=\linewidth]{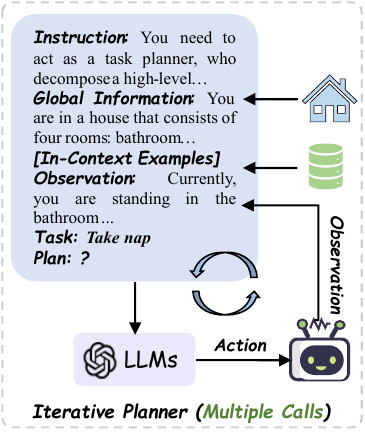}
    \end{center}
    \vspace{-0.45cm}
\caption{An overview of the traditional paradigm.}
\label{fig:iterative}
\vspace{-14pt}
\end{wrapfigure}

Task planning is a significant topic in the field of robotics, where a system is tasked with crafting a sequence of mid-level actions (skills) that enable a robot to complete complex high-level tasks~\citep{kaelblingTAMP}.
This involves a consideration of 
various factors, such as the capabilities of robots, the surrounding environment, and any constraints or uncertainties that might exist.
An emerging trend within the field of task planning is using Large Language Models (LLMs) to generate actions directly~\citep{DBLP:conf/icml/HuangAPM22,song2023llmplanner}, rather than searching within a pre-defined 
domain~\citep{Eysenbach_Salakhutdinov_Levine_2019,Xu_Martín-Martín_Huang_Zhu_Savarese_Fei-Fei_2019}.

As shown in Figure~\ref{fig:iterative}, the commonly adopted paradigm for LLM-based planning can be summarized as follows: 
\textbf{(i)} prompt an LLM to generate one action at a time; 
\textbf{(ii)} execute the generated action and then append the obtained observation to the LLM;
and 
\textbf{(iii)} generate the next action.
We categorize such approaches as \textit{\sbs}, which
enables the model to generate subsequent actions in an auto-regressive manner. 
Based on \sbs, when errors occur during action execution, existing research endeavors either re-generate actions at the current timestep~\citep{DBLP:journals/corr/abs-2211-09935,guo2023doremi} or re-generate the entire plan from the initial timestep~\citep{shinn2023reflexion},
referred to as \localreplan and \globalreplan, respectively.

All methods above have the following two drawbacks: 
\textbf{(i)} Token Inefficiency:
The expenses for a single LLM call increase proportionally with the number of tokens utilized, including both the \textit{prompt tokens} and the \textit{generated tokens}.
However, in the scenario of task planning, the \textit{prompt tokens} often consists of instructions, global information about the environment, in-context learning examples, and environmental observation~\citep{vemprala2023chatgpt} while the \textit{generated tokens} predominantly represent a concise action. 
The discrepancy in the number of tokens between \textit{prompt tokens} and \textit{generated tokens} results in the issue of token inefficiency~\citep{cheng2023batch}.
Moreover, due to the multi-step nature of a complex task (usually involving 5-20 steps), the \textit{prompt tokens} incur repeated charges, leading to even higher costs.
\textbf{(ii)} Correction Inefficiency: 
\localreplan can be viewed as a trial-and-error approach implemented at the execution-failed time step, which makes it difficult for the model to detect errors that occurred several time steps earlier. While \globalreplan can mitigate this problem by regenerating the entire plan, it may still come at the cost of increased time and token consumption.
The token and correction inefficiencies inherent in \sbs limit its applicability for large-scale inference or frequent use in everyday life.

\begin{figure}[t]
    \centering
    \includegraphics[width=1\textwidth]{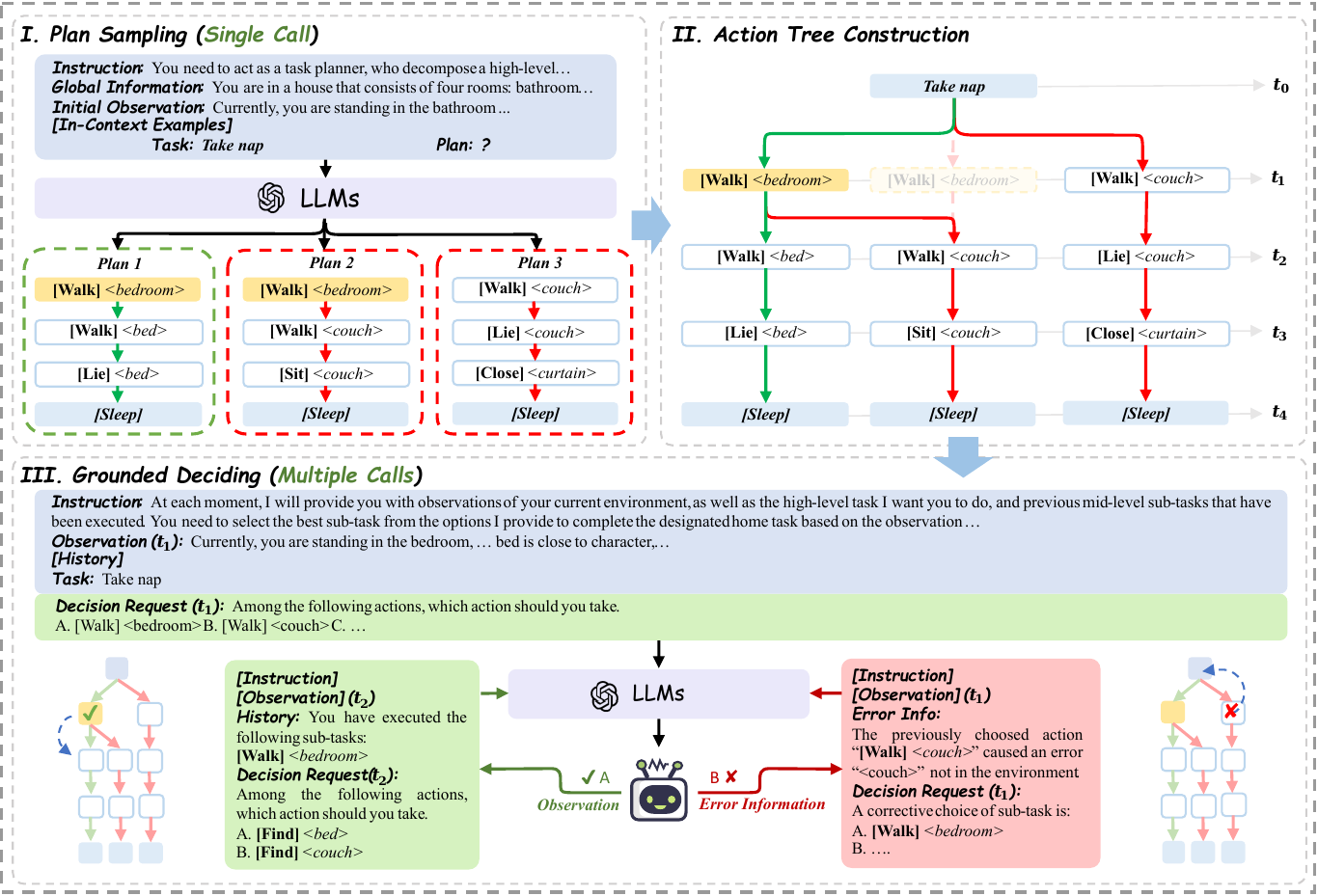}
\caption{
An overview of our \ours pipeline: 
(\uppercase\expandafter{\romannumeral 1}) prompt an LLM to sample potential plans for ``\textit{Take nap}" before execution;
(\uppercase\expandafter{\romannumeral 2}) construct an action tree to aggregate sampled plans;
(\uppercase\expandafter{\romannumeral 3}) prompt the LLM again in closed loops to reason on the action tree. 
\textit{Bottom-left}: ``[WALK] \textless bedroom\textgreater" is successfully executed. Move on to the next level. 
\textit{Bottom-right}: ``[WALK] \textless couch\textgreater" fails because the absense of ``couch". Mark the failed node as invalid, then track back and re-decide.
The complete prompt and action tree can be found in Appendix~\ref{appendix:prompt-examples} and Appendix~\ref{appendix:viz-decision-tree}, respectively.
}
\label{fig:model1}
\vspace{-0.4cm}
\end{figure}


To address the issues above while maintaining high performance, we propose \ours as illustrated in Figure~\ref{fig:model1}.
In general, \ours divides the queries to an LLM into two parts: a single plan-sampling call and multiple grounded-deciding calls to reduce the repetitive computational cost for several components in \textit{prompt tokens}. 
These two stages are bridged using a tree-like structure, which leads to more effective logical correction.
%
More specifically, \ours first prompts the LLM to sample potential task plans with its inherent commonsense (Stage \uppercase\expandafter{\romannumeral 1}).
Subsequently, an action tree is constructed to aggregate the sampled plans (Stage \uppercase\expandafter{\romannumeral 2}). 
Lastly, \ours instructs the LLM again in closed loops to reason on the action tree with the environmental observations (Stage \uppercase\expandafter{\romannumeral 3}).
In terms of token efficiency, \ours only charges once for global information about the environment and in-context examples in plan sampling. However, for \sbs, this information must be charged at each time step. 
In terms of correction efficiency, the correction process based on the action tree can be seen as an intermediate between \localreplan and \globalreplan. \ours not only reduces the likelihood of redundant decision-making at a specific time step through backtracking but also significantly reduces the time and tokens required to generate the entire plan from scratch.

We demonstrate the effectiveness of \ours framework in VirtualHome~\citep{Puig_2018_CVPR}, a simulated environment for complex household tasks. 
The experiments are conducted under two different settings: \textit{with correction} and \textit{without correction}. In \textit{with correction} setting, the model is required to modify the plan when errors occur, while in \textit{without correction} setting, the opposite is true.
The main result shows that \ours achieves state-of-the-art results in both experimental settings, surpassing the best baseline models by 1.29\% and 3.65\% in terms of success rate, respectively. 
At the same time, \ours exhibits high efficiency. 
In terms of token efficiency, \ours reduces the token cost of \sbs by 53.29\%. Furthermore, when compared to \localreplan and \globalreplan under the \textit{with correction} setting, \ours achieves even greater improvement with reductions of 74.36\% and 92.24\%, respectively.
In terms of correction efficiency, \ours reduces the number of corrections by 37.99\% and 40.52\%, respectively. 
In further analysis, we formally verify the token efficiency of \ours and derive the critical value of the number of sampled plans required for the model to possess token efficiency.
We also perform an ablation study on both plan sampling and grounded deciding, demonstrating the effectiveness of the individual components of \ours.
Finally, we provide a manual error analysis of potential areas for improvement in the model.


\section{Preliminary}
\noindent \textbf{Task and Motion Planning}
(TAMP)~\citep{kaelblingTAMP} is the process of generating a sequence of actions and robot motions to achieve a desired goal in a given environment. As is shown in Figure~\ref{fig:model1}, a high-level task description such as ``\textit{Take nap}" is decomposed into several mid-level actions. We assume the existence of a low-level controller that can execute these mid-level actions, which typically requires training using reinforcement learning (RL) methods or fine-tuning with expert data.
Task planning can be categorized into closed-loop task planning and open-loop task planning. Open-loop task planning aims to decompose a high-level task description into a mid-level plan without any feedback from the environment. 
Closed-loop task planning, on the other hand, involves continuously adjusting planning strategies through perception and feedback mechanisms to adapt to environmental changes and uncertainties during execution.
This paper focuses on closed-loop task planning, which is more suitable for task execution in dynamic and complex environments.

\noindent \textbf{Problem Setup}
We formulate the closed-loop task planning problem as a partially observable Markov decision processes (POMDPs) denoted by $\langle \states,\observation,\actions,\mathcal{T}\rangle$, which is similar to \citet{li2022pretrained}. $\states, \observation, \actions$ are sets of states, observations and actions respectively and $\mathcal{T}(s_{t+1}|s_t,a_t)$ is a transition model. In a POMDP setting, the observation $o_t$ represents a subset of the underlying state $s_t$. Let $g$ be the task, the optimal policy $\pi(a_t|g,h_t,o_t)$ must take into account not only the current observation $o_t$, but also the entire history of actions $h_t = \{a_1, \dots, a_{t-1}\}$.

\section{Model}

\subsection{Plan Sampling}
Abstract specifications often restrict task planning.
Take the ``\textit{Take nap}" task as an example, the robot needs to understand that napping can be done on a bed, and the bed is typically located in a bedroom.
Many works hold the belief that LLMs trained on large-scale data encode commonsense knowledge about the real-world~\citep{davison-etal-2019-commonsense, li-etal-2022-systematic, bian2023chatgpt}. 
Recently, several studies have investigated the integration of LLMs into task planning, which aims to address language ambiguities and provide robots with background knowledge~\citep{DBLP:conf/icml/HuangAPM22,li2022pretrained,ahn2022i}.
In contrast to these approaches, which typically use LLMs directly as policies, \ours prompts an LLM to generate prospective task plans before executing them in a specific environment. 
We consider this as a way to extract commonsense knowledge from LLM through sampling, which serves as prior knowledge for task planning.
Let $\rho_{ps}$ be the prompt for plan sampling, $g$ be the task name, the process of plan sampling can be formalized as: $\textrm{LLM}(\rho_{ps}, g) = \boldsymbol{c} = \{c_1, c_2, \dots, c_N\}$, where $N$ is a hyper-parameter which determines the number of sampled plans. Each plan candidate $c_i$ is a sequence of actions, i.e., $c_i = \{a_{it} | t = 1, \dots, m(i)\}$. $m(i)$ is the number of actions in plan $i$ and $a_{it}$ is the action of plan $i$ at time step $t$.
The prompt consists of four parts: \textit{instruction}, \textit{global information}, \textit{initial observation}, and \textit{in-context examples}. 
The \textit{instruction} provides the LLM with a clear and concise explanation of the process of task planning.
The \textit{global information} provides the LLM with background knowledge about the environment and available action space.
The \textit{initial observation} provides the LLM with an initial snapshot at the starting point of the task.
The \textit{in-context examples} are additional task plans that serve to indicate the format of the output plan and have also been proven to be helpful in enhancing performance~\citep{brown2020language}.
In Section~\ref{sec:analysis-plan-sampling}, we provide a quantitive analysis of the upper-bound on plan sampling.

\subsection{Action Tree Construction}
\label{sec:decision-tree-construction}
\begin{figure}[ht]
    \begin{center}
    \includegraphics[width=0.95\columnwidth]{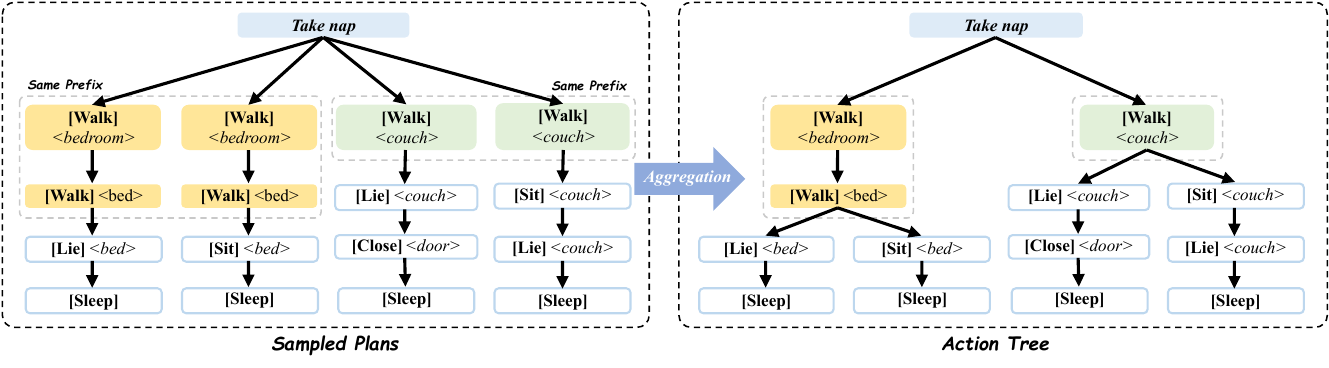}
    \end{center}
    \vspace{-0.5cm}
\caption{
The process of constructing the action tree. 
\textit{Left}: each path represents a sampled plan. \textit{Right}: plans with the same prefix are aggregated together. 
Note that although certain paths have the same action ([Sleep]), they are not aggregated together due to inconsistent prefixes.
}
\label{fig:action_tree_construction}

\end{figure}
To select an optimal plan from potential plans, an obvious approach would be to execute and test each plan in the environment. However, this approach has two drawbacks: \textbf{(i)} It is time-consuming to execute multiple plans in the environment; \textbf{(ii)} Different plans may have overlapping parts, so repeating the execution of these overlapping parts in the environment is redundant. For example, in \textit{plan 1} and \textit{plan 2} shown in Figure~\ref{fig:model1}, the first step in both plans is: ``[Walk] \textless \textit{bedroom}\textgreater".
Based on the previous analysis, we designed a structured representation that aggregates the sampled potential plans called Action Tree. As is shown in Figure~\ref{fig:action_tree_construction}, when two plans share a common prefix but differ in their actions at a specific time step, their shared prefix is aggregated into a single branch, while their differing actions form divergent paths. This process repeats iteratively until all sampled plans are organized into a complete tree structure.
The motivation behind it is to convert the filtering of the plan level into a search at the action level, thereby reducing the execution time in the environment.
an action tree with root node $r$ can be formalized as $T(\boldsymbol{c}) = (V, E)$, where $V$ and $E$ represent the sets of nodes and edges respectively.
Each node $v$ is associated with an action $a_v$ and a time step $t_v$, i.e., $v = (a_v,~t_v)$. Each edge $e$ represents a pair of adjacent actions in plan $c_i$, i.e., $E=\{e(v_1,~v_2)~|~v_1,~v_2 \in V,~v_1 = (a_{it}, t),~v_2 = (a_{i(t+1)}, t+1)\}$. The root node $r$ is not associated with any specific action, and its child nodes are the set of nodes obtained by aggregating the first action of each plan.
The construction process of the action tree is presented in Algorithm~\ref{alg:action-tree}. 

\begin{algorithm}[htbp]

\relsize{0.30} 
\caption{Action Tree Construction}
\label{alg:action-tree}
\SetKwInOut{Input}{Input}
\SetKwInOut{Output}{Output}
\SetAlgoLined
\Input{$\boldsymbol{c}$, $r$}
\Output{$r$}
\SetKwFunction{FMain}{ConstructActionTree}
\SetKwProg{Fn}{Function}{:}{}
\Fn{\FMain{$\boldsymbol{c}$, $r$}}{
    \ForAll{$c_i \in \boldsymbol{c}$}{
        $n \gets r$\;
        \For{$t=1$ \KwTo $m(i)$}{
            $cn \gets \mathrm{GetChildNode}(n,~a_{it})$\;
            \If{$cn$ is None}{
                $cn \gets \mathrm{CreateChildNode}(a_{it})$\;
                $\mathrm{AddChildNode}(n,~cn)$\;
            }
            $n \gets cn$\;
        }
    }
}
\end{algorithm}



\subsection{Grounded Deciding}

\begin{figure}[ht]
    \centering
    \includegraphics[width=1\textwidth]{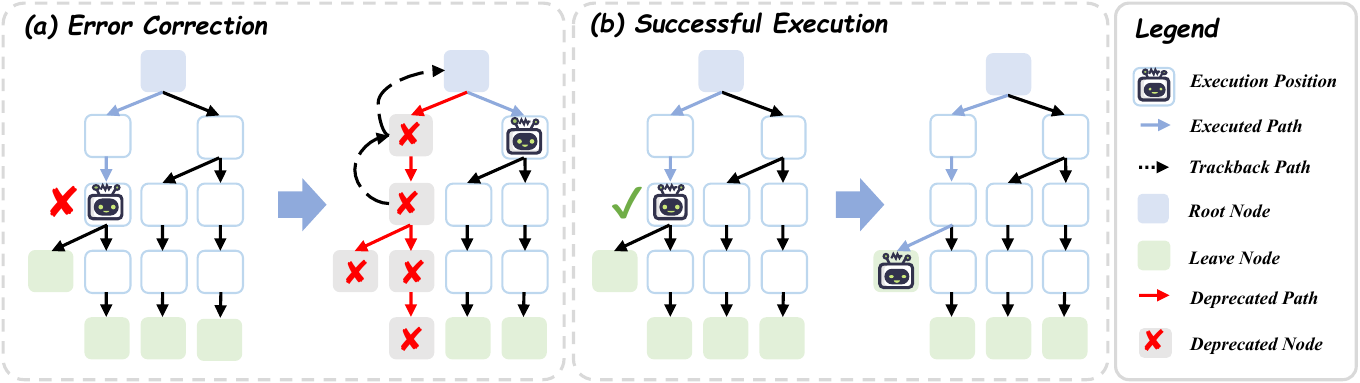}
\caption{
An overview of the process of \textit{grounded deciding}.
\textit{Left}: When an error occurs, \ours tracks back and marks the nodes along the way as invalid. Afterward, \ours makes a new decision at the previous fork node. 
\textit{Right}: After the action is successfully executed, \ours makes a decision at the current node, and then the agent moves on to the next level.
}
\label{fig:grounded-deciding}
\vspace{-0.5cm}
\end{figure}

During grounded deciding, an LLM functions as the policy $\pi(a_t | g, h_t, o_t)$. 
However, instead of sampling from the entire corpus of LLMs as the \sbs, we limit the choices to a few child nodes of the current node at time $t$ on the action tree. 
This process simulates the decision-making process of humans, who first propose several action options and then combine their current real-world observations to make decisions. 
Specifically, we provide an LLM with \textit{instruction}, \textit{observation}, and \textit{history} (the previously executed actions) as prompts, and then the LLM chooses one from the child nodes of the current node. 
Furthermore, we also designed a corresponding error correction method. 
When a chosen action fails to execute in the environment, \ours 
\textbf{(i)} marks the nodes on the subtree rooted at the failed node as invalid nodes; 
\textbf{(ii)} traces back on the action tree to find the previous valid fork node with available valid child nodes. If all the child nodes of a particular node are invalid, then the fork node should also be marked as invalid.
\textbf{(iii)} executes the inverse process of previously executed actions (e.g., the inverse of [SwitchOn] is [SwitchOff]) to recover the state of the agent;
\textbf{(iv)} re-decides at the fork node.
Error correction with grounded deciding is more effective than the commonly adopted methods presented in Section~\ref{sec:intro}. This is because the action tree serves as an important prior to completing the current task. Therefore, when an error occurs at a node on the tree, it is possible to selectively backtrack on the action tree, thus alleviating repetitive decisions at a particular time step as in \localreplan. 
Performing error correction on the action tree also relieves the need to return to the initial time step as in \globalreplan, thereby reducing time and token consumption. 
The process described above is displayed in Figure~\ref{fig:grounded-deciding}. 
Quantitive analysis of the effectiveness of error correction is presented in Section~\ref{sec:grounded-deciding-analysis}.

\section{Experimental Results}
\subsection{Experimental Setup}

\noindent \textbf{Environment.}
We conduct the experiments in the VirtualHome (VH) Environment~\citep{Puig_2018_CVPR}, a simulation platform for household tasks. Each scene in every VH environment contains hundreds of objects. These objects may possess individual properties, and there may also be relationships between different objects. There are 28 different action types in VH, which are listed in Appendix~\ref{appendix:action-types}. 
The task-relevant goal conditions refer to a set of specific states of objects or predicates between objects. For example, a goal condition for \textit{Turn on TV} would be \textit{On(TV)}, while a goal condition for \textit{Sit} would be \textit{On(character, chair)}. 

\noindent \textbf{Dataset.}
We constructed a dataset consisting of 4 VH scenes and 35 unique VH tasks. Each task includes a task name, goal conditions, and a gold plan.
We started by annotating goal conditions for each task from \textit{ActivityPrograms} knowledge base by \cite{Puig_2018_CVPR} via executing the programs. And then, we applied simple heuristics to filter the low-quality annotations in the dataset: \textbf{(i)} the length of the plan is less than 3; \textbf{(ii)} the execution of the program fails.
To highlight the necessity of grounding LLMs in the real environment which has variation in the objects and preconditions, we replicated the annotation above process across 4 distinct scenes provided in VirtualHome, ultimately yielding 71 annotated tasks. 
We denote the 4 distinct scenes as ENV-\{1, 2, 3, 4\}.
Then, we hired two CS-majored graduate students to conduct manual quality control to ensure that the task descriptions were in line with their corresponding goal conditions and programs. We eliminate cases that do not meet the alignment criteria or were originally annotated with errors, resulting in a high-quality dataset comprising 35 tasks. To double-check the quality of the dataset, we also study the agreement between annotators. The results indicated ``almost perfect agreement" with Fleiss Kappa~\citep{10.2307/2529310} scores of 0.88.

\noindent \textbf{Evaluation Metrics.}
We use four metrics to evaluate the performance of different methods: executability (\exec), success rate (SR),  goal conditions recall (GCR), and the financial expenditure for evaluation (\cost). \exec refers to whether the plan can be executed in the given environment, regardless of its relevance to the task. 
GCR is calculated by taking the difference between the ground truth goal conditions and the goal conditions that were achieved with the generated plan and then dividing this difference by the total number of goal conditions. SR measures whether all goal conditions are fulfilled, i.e., $SR=1$ only when $GCR=1$. \cost is used to evaluate the token efficiency of different methods, which is calculated based on the pricing provided by OpenAI.\footnote{\url{https://openai.com/pricing}}
For evaluation with error correction, we use \nocorrection to represent the number of error corrections of each method. \nocorrection does not directly measure performance but rather evaluates how effectively different models can correct errors.

\noindent \textbf{Baselines.} 
For experiments without error correction, we compare our method to two strong published LLM-based task planning methods with OpenAI APIs, including: \textbf{(i)} \zeroshot~\citep{DBLP:conf/icml/HuangAPM22}; \textbf{(ii)} \progprompt~\citep{singh2022progprompt}. 
Furthermore, we also implement the \sbs method discussed in Section \ref{sec:intro} as a baseline model.
For experiments with error correction, we enhance the \sbs method with the two re-planning methods: \localreplan and \globalreplan, and consider them as the baseline models. 
More implementation details and an introduction to each baseline model can be found in Appendix \ref{appendix:baseline-models-details}.


\noindent \textbf{Implementation Details.}
We use the OpenAI GPT-3.5 (text-davinci-003) API \footnote{\url{https://openai.com/}} model as a LLM backbone in our experiments for all evaluated methods. The cost of this model is 0.02\$ per 1000 tokens.
The prompt for \ours and \sbs was designed with the principles proposed in \citet{vemprala2023chatgpt}, and examples can be found in Appendix~\ref{appendix:prompt-examples}.
We take 4 representative tasks from the dataset as in-context learning exemplars and the rest as the validation set. The examples are fixed to be: ``\textit{Watch TV}", ``\textit{Turn on light}", ``\textit{Go to sleep}", and ``\textit{Brush teeth}".
To sample diverse plans, we applied a temperature of 0.8 and a top-p value of 0.95. We heuristically set the number of samplings $N \in \{25, 50\}$. 
During grounded deciding, we set the temperature to 0.7, top-p to 1.0, and sampling parameter n to 20. Additionally, we utilize a majority vote to obtain the final option in order to alleviate format errors in the output of LLMs.
The maximum number of error corrections is set to 10 for all evaluated approaches.

\subsection{Main Results}
\label{sec:main-results}

\begin{table*}[htbp]
  \centering
    \caption{Performance of different methods on Virtual Home. \textit{w/o correction} means that during the plan execution, there is no allowance for retrying failed actions. While \textit{with correction} implies the opposite. The reported evaluation metrics are the average of 3 independent runs across the 4 scenes. 
    }
    \vspace{-8pt}
    \begin{tabular}{lccccc}
    \toprule
          & \textbf{\exec} $\uparrow$ & \textbf{SR} $\uparrow$ & \textbf{GCR} $\uparrow$ & \textbf{\cost} $\downarrow$ & \textbf{\nocorrection} $\downarrow$ \\
    \midrule
    \multicolumn{6}{l}{\textbf{\textit{w/o correction}}} \\
    \small{\zeroshot} & 16.49$\pm$3.08 & 1.07$\pm$0.76 & 1.52$\pm$0.75 & 1.36$\pm$0.09 & N/A \\
    \small{\progprompt} & 35.04$\pm$3.98 & 12.54$\pm$2.20 & 19.99$\pm$2.83 & \textbf{1.25}$\pm$0.55 & N/A \\
    \small{\sbs}  & 44.54$\pm$6.09 & 27.04$\pm$4.65 & 33.25$\pm$5.32 & 5.12$\pm$0.14 & N/A \\
    \rowcolor{gray!15}
    \small{$\ours_{N=25}$} & \textbf{55.74}$\pm$0.92 & \textbf{28.33}$\pm$1.18 & \textbf{39.96}$\pm$0.16 & 2.39$\pm$0.44 & N/A \\
    \rowcolor{gray!15}
    \small{$\ours_{N=50}$} & 49.01$\pm$5.67 & 28.14$\pm$2.45 & 35.84$\pm$4.20 & 3.48$\pm$0.04 & N/A \\
    \midrule
    \multicolumn{6}{l}{\textbf{\textit{with correction}}} \\
    \small{\localreplan} & 79.66$\pm$2.33 & 37.46$\pm$1.71 & 51.9$\pm$0.15 & 12.88$\pm$0.17 & 3.29$\pm$0.46 \\
    \small{\globalreplan} & 82.09$\pm$1.32 & 37.93$\pm$1.22 & 52.46$\pm$0.86 & 42.55$\pm$0.09 & 3.43$\pm$0.15 \\
    \rowcolor{gray!15}
    \small{$\ours_{N=25}$} & \textbf{89.13}$\pm$0.17 & 35.30$\pm$1.78 & 56.65$\pm$1.09 & \textbf{3.30}$\pm$0.01 & \textbf{1.85}$\pm$0.05 \\
    \rowcolor{gray!15}
    \small{$\ours_{N=50}$} & 88.26$\pm$2.47 & \textbf{41.58}$\pm$3.20 & \textbf{59.55}$\pm$3.20 & 4.54$\pm$0.16 & 2.04$\pm$0.26 \\
    \bottomrule
    \end{tabular}%
  \label{tab:main-results}%
\vspace{-8pt}
\end{table*}

Based on the results presented in Table~\ref{tab:main-results}, several advantages of \ours can be derived:
\textbf{(i)} \ours outperforms listed baseline systems, surpassing the previous state-of-the-art by absolute 11.2\% and 7.04\% on Executability, 6.71\% and 7.29\% on \gcr and 1.29\% and 3.65\% on \sr under both experimental settings respectively.
This experimental observation demonstrates that reframing the LLM-based planning pipeline does not compromise its performance.
\textbf{(ii)} \ours has a significant advantage in token efficiency. In \textit{without correction} setting, \ours reduces the cost of \sbs by 53.29\%. In \textit{with correction} setting, the token cost is further reduced by 74.36\% and 92.24\%, respectively, compared to \localreplan and \globalreplan.
\textbf{(iii)} \ours also demonstrates high correction efficiency, resulting in a reduction of the number of action-retry times for \localreplan and \globalreplan by 37.99\% and 40.52\%, respectively.
A reduced amount of corrections also contributes to a decrease in token consumption.

Note that, while not having a token efficiency advantage compared to \zeroshot and \progprompt, \ours significantly outperforms these methods in terms of performance by 27.26\% and 15.79\% on \sr respectively. 
It is also worth noting that increasing the hyper-parameter $N$ does not result in consistently improved performance. This experimental phenomenon will be further discussed in Section~\ref{sec:analysis-plan-sampling}.

\section{Analysis}

\subsection{Token Efficiency}
\label{sec:token-efficiency-analysis}
In Section~\ref{sec:main-results}, the quantitative analysis has demonstrated that \ours consumes fewer tokens compared to \sbs.
In this section, we will further provide a specific formulation to demonstrate this point.
The number of tokens required for an LLM API call typically includes two parts: \textit{prompt tokens} and \textit{generated tokens}. Let $\rho$ and $\varphi$ represent the prompt tokens and generated tokens. Let $ps,~gd,~ip$ stand for plan sampling, grounded deciding, and \sbs, respectively. Normally, we have $\rho_{ip}\approx \rho_{ps}+\rho_{gd}$. That is because, as shown in Figure~\ref{fig:model1} and Figure~\ref{fig:iterative}, the prompt for plan sampling typically includes global information and in-context examples, while the prompt for grounded deciding includes observation and history. These types of information usually need to be included in every step of \sbs.
Assuming that the number of tokens for each action type $|a|$ is the same and the total number of steps $M$ is the same for each generated plan. The hyper-parameter number of sampling is $N$ for plan sampling and grounded decoding and is 1 for \sbs. Based on the given information, we have $\varphi_{ps} = MN|a|$, $\varphi_{gd} = N$ and $\varphi_{ip}=|a|$.
The consumed tokens $\mu_{ours}$ and $\mu_{ip}$ can be calculated as follows: $\mu_{ours}=\rho_{ps} + \varphi_{ps} + M \cdot (\rho_{gd} + \varphi_{gd})$ and $\mu_{ip}=M \cdot (\rho_{ip} + \varphi_{ip})$. 
Based on the above formula, we can determine the boundary conditions for $N$ that satisfy the inequality $\mu_{ours} < \mu_{ip}$ as follows: $N < \frac{1-1/M}{1+1/|a|} \cdot \frac{\rho_{ps}}{|a|} + \frac{|a|}{|a| + 1}$. And we have $\rho_{ps} \gg |a|$, since the prompt of plan sampling may contain thousands of tokens and an action only contains a few tokens. 
We use the average number of tokens for all action types to estimate $|a|$ and the average length of all gold plans to estimate $M$. As a result, we obtain the critical value of $N$ in our experiment as follows: $N < 197.72$.
Detailed derivation can be found in Appendix~\ref{appendix:token-efficiency}.
In conclusion, our model exhibits a remarkably high token efficiency, especially in scenarios where $N$ is not particularly high.

\subsection{Plan Sampling}
\label{sec:analysis-plan-sampling}

\begin{figure}[htbp]
  \begin{minipage}[b]{0.44\textwidth}
    \centering
    \includegraphics[width=0.98\linewidth]{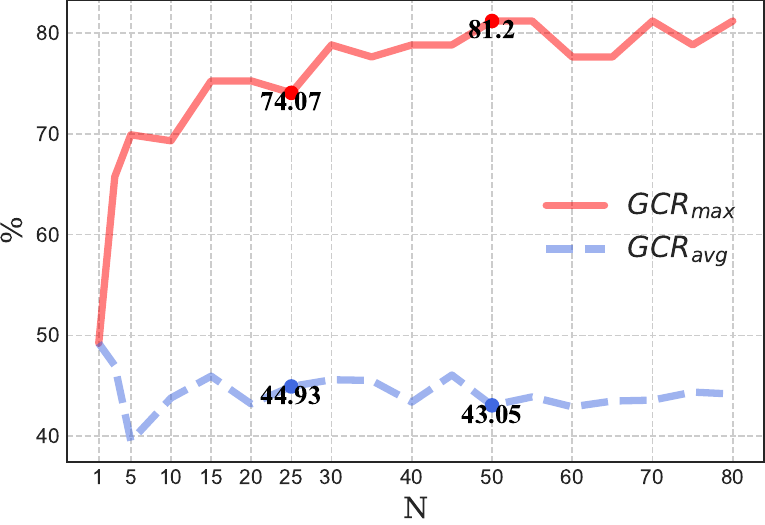}
    \vspace{-6pt}
    \caption{Maximum and average GCR for all sampled plans. The x-axis of the annotated coordinate points represents the chosen $N$ for the main experiments.}
    \label{fig:upper-bound-ps}
  \end{minipage}%
  \hfill
  \begin{minipage}[b]{0.52\textwidth}
    \centering
    \small
    \captionsetup{type=table}
    \begin{tabular}{lccc}
    \toprule
          & \textbf{\exec} & \textbf{SR} & \textbf{GCR} \\
    \midrule
    \multicolumn{4}{l}{\textbf{\textit{w/o correction}}} \\
    \multicolumn{1}{l}{\small{$\ours_{N=25}$}} & 55.74 & 28.33 & 38.96 \\
        $\dagger $ with oracle $\uparrow$ & \cellcolor[rgb]{ .988,  .847,  .859} 7.16 & \cellcolor[rgb]{ .984,  .788,  .796} 9.84 & \cellcolor[rgb]{ .984,  .816,  .827} 8.5 \\
    \multicolumn{1}{l}{\small{$\ours_{N=50}$}} & 49.01 & 28.14 & 35.84 \\
        $\dagger$ with oracle $\uparrow$ & \cellcolor[rgb]{ .988,  .929,  .941} 3.41 & \cellcolor[rgb]{ .988,  .859,  .871} 6.54 & \cellcolor[rgb]{ .988,  .898,  .91} 4.78 \\
    \midrule
    \multicolumn{1}{l}{\textbf{\textit{with correction}}} & \multicolumn{3}{l}{} \\
    \multicolumn{1}{l}{\small{$\ours_{N=25}$}} & 89.13 & 35.3  & 56.65 \\
        $\dagger$ with oracle $\uparrow$ & \cellcolor[rgb]{ .984,  .82,  .827} 8.45 & \cellcolor[rgb]{ .973,  .412,  .42} 26.8 & \cellcolor[rgb]{ .98,  .569,  .576} 19.76 \\
    \multicolumn{1}{l}{\small{$\ours_{N=50}$}} & 88.26 & 41.58 & 59.55 \\
        $\dagger$ with oracle $\uparrow$ & \cellcolor[rgb]{ .988,  .851,  .863} 6.9 & \cellcolor[rgb]{ .984,  .773,  .78} 10.57 & \cellcolor[rgb]{ .984,  .839,  .851} 7.47 \\
    \bottomrule
    \end{tabular}%
    \vspace{-6pt}
    \captionof{table}{Ablation study on grounded deciding. $\dagger$ represents the performance improvement after adding a gold plan to action tree construction.
    }
    \label{tab:ablation-gd}
  \end{minipage}
  \vspace{-9pt}
\end{figure}


Since grounded deciding fundamentally involves selecting from the sampled plans, the upper limit of our \ours is determined by plan sampling. 
We propose two additional metrics to study the upper limit of plan sampling:
\textbf{(i)} the maximum GCR for all generated plans, i.e., {\small $GCR_{max}(\boldsymbol{c}) = \max\limits_{i=1}^{N} (GCR(c_i))$};
\textbf{(ii)} the average GCR for all generated plans, i.e., {\small $GCR_{avg}(\boldsymbol{c}) = \frac{1}{N} \sum_{i=1}^{N} (GCR(c_i))$}.
$GCR_{max}$ represents the upper limit of the performance of \ours. In other words, the model can only succeed if there is a ``\textit{correct}" plan among the sampled plans. $GCR_{avg}$ reflects the proportion of ``\textit{correct}" plans to sampled plans. When $GCR_{avg}$ is low, it undoubtedly poses greater challenges for grounded deciding.
Some conclusions can be drawn from Figure~\ref{fig:upper-bound-ps}:
\textbf{(i)} The maximum value of $GCR_{max}$ being 81.2\% indicates that plan sampling is effective.
\textbf{(ii)} As $N$ increases, there is a noticeable increase in $GCR_{max}$, but it eventually reaches a threshold. 
Therefore, a large value of $N$ will lead to increased token consumption without necessarily improving the performance limit. 
When applying \ours, it is essential to choose an appropriate value of $N$ that balances token assumption and model performance.
\textbf{(iii)} $GCR_{avg}$ does not consistently increase with an increased $N$. This implies that as $N$ becomes larger, the proportion of ``\textit{correct}" plans to sampled plans may not necessarily increase. 

\subsection{Grounded Deciding}
\label{sec:grounded-deciding-analysis}

To investigate the effectiveness of grounded deciding, we conducted ablation experiments. We incorporated the gold plan for each task into the construction of the action tree.
As is shown in Table~\ref{tab:ablation-gd}, after incorporating the gold plan, there was a significant improvement in performance.
Additionally, there was also a decrease in the number of error corrections. For $\ours_{N=25}$, the number decreased from 1.85 to 1.21, and for $\ours_{N=50}$, it decreased from 2.04 to 1.39. 
The quantitive experimental results presented above demonstrate the effectiveness of grounded deciding.
Another noteworthy experimental phenomenon is that the improvement in performance for $\ours_{N=25}$ was greater than that for $\ours_{N=50}$. This further validates the conclusion we drew in Section~\ref{sec:analysis-plan-sampling}: when the number of plans increases, but the proportion of correct plans decreases, the performance may be negatively impacted.

\subsection{Error Analysis}


\begin{table}[b]
  \centering
  \small
  \caption{Distribution of error types of the $\ours_{N=25}$ \textit{w/o correction} model.}
  \vspace{-6pt}
    \begin{tabular}{lp{6cm}l}
    \toprule
    Error Type & Explanation & \multicolumn{1}{l}{Proportion(\%)} \\
    \midrule
    Missing Correct Plans & Plan sampling did not yield correct plans & \textbf{54.5\%} \\
    ~~~~Environment Misunderstanding & Misunderstandings on actions or objects & ~~~~18.2\% \\
    ~~~~Incomplete Plan & The absence of essential steps & ~~~~18.2\% \\
    ~~~~Illogical Error & The generated plan is logically incorrect & ~~~~13.6\% \\
    ~~~~Semantically Correct & Execution failed but semantically correct & ~~~~9.1\% \\
    Grounded Deciding Error & Execution error during grounded deciding & \textbf{45.5\%} \\
    ~~~~Incorrect Deciding & Incorrect decisions at specific nodes & ~~~~31.8\% \\
    ~~~~Semantically Correct & Execution failed but semantically correct & ~~~~13.7\% \\
    \bottomrule
    \end{tabular}%
  \label{tab:error-types}%
\end{table}%

We categorize error types into two distinct classifications: \textbf{(i)} Missing Correct Plan; \textbf{(ii)} Grounded Deciding Error. 
As is listed in Table~\ref{tab:error-types}, the majority of errors are attributed to the missing correct plans (54.5\%). Therefore, despite the ability of plan sampling to achieve relatively high $GCR_{max}$ as is discussed in Section~\ref{sec:analysis-plan-sampling}, it still serves as a bottleneck for our model to some extent. 
%
Furthermore, a considerable portion of the errors occurred due to mistakes made by LLM during grounded deciding (45.5\%).
%
We also provide a qualitative analysis of each error type in Appendix~\ref{appendix:error-types}.


\section{Related Works}
\noindent \textbf{Task Planning with Large Language Models.}
We categorize the mainstream methods in the task planning domain into two groups: search-based methods~\citep{Jiang_Zhang_Khandelwal_Stone_2018,Garrett_Lozano-Pérez_Kaelbling_2018} and generate-based methods~\citep{song2023llmplanner,wu2023plan,ding2023task,mu2023embodiedgpt}. LLMs trained on a large-scale corpus contain a vast amount of commonsense knowledge for task planning~\citep{pallagani2023understanding,sun2023pearl,sun2023adaplanner}. 
Thanks to this advancement, generate-based methods have gradually become a hot topic of research in recent years. 
When considering the utilization of LLM, some works directly generate the entire plan without executing in the environment~\citep{singh2022progprompt, liang2023code, wu2023embodied, zeng2023socratic, lin2023grounded, yang2023foundation}. While these models possess token efficiency, they are unable to modify the plan when encountering errors dynamically.
Another line of works has adopted the paradigm presented in Section~\ref{sec:intro} to generate actions iteratively~\citep{vemprala2023chatgpt, yao2022react, DBLP:conf/icml/HuangAPM22, huang2022inner, shinn2023reflexion}, which is more flexible for error correction, human interaction and the grounding of environment.  
Works like~\citet{carta2023grounding,huang2023grounded,ahn2022i} involve the use of implicit representations of LLM.
In contrast to these works, our study concentrates on Black-box LLMs, which are utilized in a manner more frequently by researchers and industry, as they provide only input and output without any additional information.

\noindent \textbf{Tree-based Modeling for the Output of Large Language Models.}
\citet{yao2023tree,long2023large} both propose an alternative for chain-of-thought, called ``tree-of-thought", for problem-solving.
These studies do not involve the interaction between inner steps in the tree and the environment but rather focus on reasoning tasks.
Considering the robotic area, \citet{cao2023robot} leverages LLMs for automatic behavior-tree-based task generation.
\citet{zhao2023large,hao2023reasoning} propose using an LLM as a world model to assist planning algorithms such as Monte Carlo Tree Search (MCTS).
However, \ours samples diverse paths once and aggregates the paths into an action tree rather than requiring multiple calls to LLM like the aforementioned studies.
This approach offers advantages in terms of both run-time efficiency and token efficiency.

\noindent \textbf{Generate then Select.}
From another perspective, grounded deciding selects a prediction from the sampled potential plans.
Hence, \ours follows the paradigm of \textit{generate then select}, which is commonly adopted to optimize the output of LLMs.
Some models~\citep{glass2022re2g,suzgun2022promptandrerank,wang2023describe,gu2023dont} use external controllers to re-rank the generations.
In~\citet{wang2023selfconsistency}, the best answer is selected from multiple generations of an LLM through a majority vote. 
\citet{logeswaran-etal-2022-shot} proposes to incorporate the state information from the environment to re-rank the generated plans.
Unlike these works, instead of selecting at the level of entire generation, we use action trees to perform more fine-grained selection (action-level).

\noindent \textbf{Efficient Inference with Large Language Models.} 
Most previous works suggest modifying the architecture of transformer or decoding strategy to achieve efficient inference~\citep{wang2020linformer,katharopoulos2020transformers,leviathan2023fast,chen2023accelerating}.
~\citet{cheng2023batch} propose a batch prompting method to reduce the frequency of invoking LLMs.
~\citet{lin2023swiftsage} achieve efficient inference with LLMs by incorporating a small LM fine-tuned on oracle trajectories.
\ours differs from previous studies by simply reframing the process of LLM planning to alleviate repeated token consumption without the need for additional training.

\section{Conclusion}
In this paper, we have introduced \ours, a novel framework for task planning with LLMs. 
The motivation behind \ours is to address the inefficiencies of the commonly adopted paradigm while still achieving high performance.
Through extensive experiments in the VirtualHome environment, we have demonstrated that \ours outperforms other strong baselines and achieves state-of-the-art performance. We have also shown that our framework is highly efficient in terms of token consumption and error correction.
To gain a deeper understanding of our framework, we have conducted several studies analyzing its performance gains and identifying potential bottlenecks. Furthermore, we have performed a qualitative error analysis to identify areas where the model may fail.
Overall, we believe that \ours represents a new paradigm for LLM-based task planning that strikes a balance between efficiency and performance. We hope that our work will inspire further research and the development of more efficient task-planning methods.

\section{Ethics Statements}

We build the dataset based on the \textit{ActivityPrograms} knowledge base by \cite{Puig_2018_CVPR}, which is under the MIT license.
Our approach has no ethical or social issues on its own, except those inherited from large language models.

\section{Acknowledgments}

This paper is partially supported by the National Key R\&D Program of China No.2022ZD0161000 and the General Research Fund of Hong Kong No.17200622. 

\bibliography{iclr2024_conference}

\begin{thebibliography}{52}
\providecommand{\natexlab}[1]{#1}
\providecommand{\url}[1]{\texttt{#1}}
\expandafter\ifx\csname urlstyle\endcsname\relax
  \providecommand{\doi}[1]{doi: #1}\else
  \providecommand{\doi}{doi: \begingroup \urlstyle{rm}\Url}\fi

\bibitem[Ahn et~al.(2022)Ahn, Brohan, Brown, Chebotar, Cortes, David, Finn, Fu, Gopalakrishnan, Hausman, Herzog, Ho, Hsu, Ibarz, Ichter, Irpan, Jang, Ruano, Jeffrey, Jesmonth, Joshi, Julian, Kalashnikov, Kuang, Lee, Levine, Lu, Luu, Parada, Pastor, Quiambao, Rao, Rettinghouse, Reyes, Sermanet, Sievers, Tan, Toshev, Vanhoucke, Xia, Xiao, Xu, Xu, Yan, and Zeng]{ahn2022i}
Michael Ahn, Anthony Brohan, Noah Brown, Yevgen Chebotar, Omar Cortes, Byron David, Chelsea Finn, Chuyuan Fu, Keerthana Gopalakrishnan, Karol Hausman, Alex Herzog, Daniel Ho, Jasmine Hsu, Julian Ibarz, Brian Ichter, Alex Irpan, Eric Jang, Rosario~Jauregui Ruano, Kyle Jeffrey, Sally Jesmonth, Nikhil~J Joshi, Ryan Julian, Dmitry Kalashnikov, Yuheng Kuang, Kuang-Huei Lee, Sergey Levine, Yao Lu, Linda Luu, Carolina Parada, Peter Pastor, Jornell Quiambao, Kanishka Rao, Jarek Rettinghouse, Diego Reyes, Pierre Sermanet, Nicolas Sievers, Clayton Tan, Alexander Toshev, Vincent Vanhoucke, Fei Xia, Ted Xiao, Peng Xu, Sichun Xu, Mengyuan Yan, and Andy Zeng.
\newblock Do as i can, not as i say: Grounding language in robotic affordances, 2022.

\bibitem[Bian et~al.(2023)Bian, Han, Sun, Lin, Lu, and He]{bian2023chatgpt}
Ning Bian, Xianpei Han, Le~Sun, Hongyu Lin, Yaojie Lu, and Ben He.
\newblock Chatgpt is a knowledgeable but inexperienced solver: An investigation of commonsense problem in large language models, 2023.

\bibitem[Brown et~al.(2020)Brown, Mann, Ryder, Subbiah, Kaplan, Dhariwal, Neelakantan, Shyam, Sastry, Askell, Agarwal, Herbert-Voss, Krueger, Henighan, Child, Ramesh, Ziegler, Wu, Winter, Hesse, Chen, Sigler, Litwin, Gray, Chess, Clark, Berner, McCandlish, Radford, Sutskever, and Amodei]{brown2020language}
Tom~B. Brown, Benjamin Mann, Nick Ryder, Melanie Subbiah, Jared Kaplan, Prafulla Dhariwal, Arvind Neelakantan, Pranav Shyam, Girish Sastry, Amanda Askell, Sandhini Agarwal, Ariel Herbert-Voss, Gretchen Krueger, Tom Henighan, Rewon Child, Aditya Ramesh, Daniel~M. Ziegler, Jeffrey Wu, Clemens Winter, Christopher Hesse, Mark Chen, Eric Sigler, Mateusz Litwin, Scott Gray, Benjamin Chess, Jack Clark, Christopher Berner, Sam McCandlish, Alec Radford, Ilya Sutskever, and Dario Amodei.
\newblock Language models are few-shot learners, 2020.

\bibitem[Cao \& Lee(2023)Cao and Lee]{cao2023robot}
Yue Cao and C.~S.~George Lee.
\newblock Robot behavior-tree-based task generation with large language models, 2023.

\bibitem[Carta et~al.(2023)Carta, Romac, Wolf, Lamprier, Sigaud, and Oudeyer]{carta2023grounding}
Thomas Carta, Clément Romac, Thomas Wolf, Sylvain Lamprier, Olivier Sigaud, and Pierre-Yves Oudeyer.
\newblock Grounding large language models in interactive environments with online reinforcement learning, 2023.

\bibitem[Chen et~al.(2023)Chen, Borgeaud, Irving, Lespiau, Sifre, and Jumper]{chen2023accelerating}
Charlie Chen, Sebastian Borgeaud, Geoffrey Irving, Jean-Baptiste Lespiau, Laurent Sifre, and John Jumper.
\newblock Accelerating large language model decoding with speculative sampling, 2023.

\bibitem[Cheng et~al.(2023)Cheng, Kasai, and Yu]{cheng2023batch}
Zhoujun Cheng, Jungo Kasai, and Tao Yu.
\newblock Batch prompting: Efficient inference with large language model apis, 2023.

\bibitem[Davison et~al.(2019)Davison, Feldman, and Rush]{davison-etal-2019-commonsense}
Joe Davison, Joshua Feldman, and Alexander Rush.
\newblock Commonsense knowledge mining from pretrained models.
\newblock In \emph{Proceedings of the 2019 Conference on Empirical Methods in Natural Language Processing and the 9th International Joint Conference on Natural Language Processing (EMNLP-IJCNLP)}, pp.\  1173--1178, Hong Kong, China, November 2019. Association for Computational Linguistics.
\newblock \doi{10.18653/v1/D19-1109}.
\newblock URL \url{https://aclanthology.org/D19-1109}.

\bibitem[Ding et~al.(2023)Ding, Zhang, Paxton, and Zhang]{ding2023task}
Yan Ding, Xiaohan Zhang, Chris Paxton, and Shiqi Zhang.
\newblock Task and motion planning with large language models for object rearrangement, 2023.

\bibitem[Eysenbach et~al.(2019)Eysenbach, Salakhutdinov, and Levine]{Eysenbach_Salakhutdinov_Levine_2019}
Benjamin Eysenbach, Ruslan Salakhutdinov, and Sergey Levine.
\newblock Search on the replay buffer: Bridging planning and reinforcement learning.
\newblock \emph{arXiv: Artificial Intelligence,arXiv: Artificial Intelligence}, Jun 2019.

\bibitem[Garrett et~al.(2018)Garrett, Lozano-Pérez, and Kaelbling]{Garrett_Lozano-Pérez_Kaelbling_2018}
CaelanReed Garrett, Tomás Lozano-Pérez, and LesliePack Kaelbling.
\newblock Pddlstream: Integrating symbolic planners and blackbox samplers via optimistic adaptive planning.
\newblock \emph{arXiv: Artificial Intelligence,arXiv: Artificial Intelligence}, Feb 2018.

\bibitem[Glass et~al.(2022)Glass, Rossiello, Chowdhury, Naik, Cai, and Gliozzo]{glass2022re2g}
Michael Glass, Gaetano Rossiello, Md~Faisal~Mahbub Chowdhury, Ankita~Rajaram Naik, Pengshan Cai, and Alfio Gliozzo.
\newblock Re2g: Retrieve, rerank, generate, 2022.

\bibitem[Gu et~al.(2023)Gu, Deng, and Su]{gu2023dont}
Yu~Gu, Xiang Deng, and Yu~Su.
\newblock Don't generate, discriminate: A proposal for grounding language models to real-world environments, 2023.

\bibitem[Guo et~al.(2023)Guo, Wang, Zha, Jiang, and Chen]{guo2023doremi}
Yanjiang Guo, Yen-Jen Wang, Lihan Zha, Zheyuan Jiang, and Jianyu Chen.
\newblock Doremi: Grounding language model by detecting and recovering from plan-execution misalignment, 2023.

\bibitem[Hao et~al.(2023)Hao, Gu, Ma, Hong, Wang, Wang, and Hu]{hao2023reasoning}
Shibo Hao, Yi~Gu, Haodi Ma, Joshua~Jiahua Hong, Zhen Wang, Daisy~Zhe Wang, and Zhiting Hu.
\newblock Reasoning with language model is planning with world model.
\newblock \emph{arXiv preprint arXiv:2305.14992}, 2023.

\bibitem[Huang et~al.(2022{\natexlab{a}})Huang, Abbeel, Pathak, and Mordatch]{DBLP:conf/icml/HuangAPM22}
Wenlong Huang, Pieter Abbeel, Deepak Pathak, and Igor Mordatch.
\newblock Language models as zero-shot planners: Extracting actionable knowledge for embodied agents.
\newblock In Kamalika Chaudhuri, Stefanie Jegelka, Le~Song, Csaba Szepesv{\'{a}}ri, Gang Niu, and Sivan Sabato (eds.), \emph{International Conference on Machine Learning, {ICML} 2022, 17-23 July 2022, Baltimore, Maryland, {USA}}, volume 162 of \emph{Proceedings of Machine Learning Research}, pp.\  9118--9147. {PMLR}, 2022{\natexlab{a}}.
\newblock URL \url{https://proceedings.mlr.press/v162/huang22a.html}.

\bibitem[Huang et~al.(2022{\natexlab{b}})Huang, Xia, Xiao, Chan, Liang, Florence, Zeng, Tompson, Mordatch, Chebotar, Sermanet, Brown, Jackson, Luu, Levine, Hausman, and Ichter]{huang2022inner}
Wenlong Huang, Fei Xia, Ted Xiao, Harris Chan, Jacky Liang, Pete Florence, Andy Zeng, Jonathan Tompson, Igor Mordatch, Yevgen Chebotar, Pierre Sermanet, Noah Brown, Tomas Jackson, Linda Luu, Sergey Levine, Karol Hausman, and Brian Ichter.
\newblock Inner monologue: Embodied reasoning through planning with language models, 2022{\natexlab{b}}.

\bibitem[Huang et~al.(2023)Huang, Xia, Shah, Driess, Zeng, Lu, Florence, Mordatch, Levine, Hausman, and Ichter]{huang2023grounded}
Wenlong Huang, Fei Xia, Dhruv Shah, Danny Driess, Andy Zeng, Yao Lu, Pete Florence, Igor Mordatch, Sergey Levine, Karol Hausman, and Brian Ichter.
\newblock Grounded decoding: Guiding text generation with grounded models for robot control, 2023.

\bibitem[Jiang et~al.(2018)Jiang, Zhang, Khandelwal, and Stone]{Jiang_Zhang_Khandelwal_Stone_2018}
Yuqian Jiang, Shiqi Zhang, Piyush Khandelwal, and Peter Stone.
\newblock Task planning in robotics: an empirical comparison of pddl-based and asp-based systems.
\newblock \emph{Cornell University - arXiv,Cornell University - arXiv}, Apr 2018.

\bibitem[Kaelbling \& Lozano-Pérez(2011)Kaelbling and Lozano-Pérez]{kaelblingTAMP}
Leslie~Pack Kaelbling and Tomás Lozano-Pérez.
\newblock Hierarchical task and motion planning in the now.
\newblock In \emph{2011 IEEE International Conference on Robotics and Automation}, pp.\  1470--1477, 2011.
\newblock \doi{10.1109/ICRA.2011.5980391}.

\bibitem[Katharopoulos et~al.(2020)Katharopoulos, Vyas, Pappas, and Fleuret]{katharopoulos2020transformers}
Angelos Katharopoulos, Apoorv Vyas, Nikolaos Pappas, and François Fleuret.
\newblock Transformers are rnns: Fast autoregressive transformers with linear attention, 2020.

\bibitem[Landis \& Koch(1977)Landis and Koch]{10.2307/2529310}
J.~Richard Landis and Gary~G. Koch.
\newblock The measurement of observer agreement for categorical data.
\newblock \emph{Biometrics}, 33\penalty0 (1):\penalty0 159--174, 1977.
\newblock ISSN 0006341X, 15410420.
\newblock URL \url{http://www.jstor.org/stable/2529310}.

\bibitem[Leviathan et~al.(2023)Leviathan, Kalman, and Matias]{leviathan2023fast}
Yaniv Leviathan, Matan Kalman, and Yossi Matias.
\newblock Fast inference from transformers via speculative decoding, 2023.

\bibitem[Li et~al.(2022{\natexlab{a}})Li, Puig, Paxton, Du, Wang, Fan, Chen, Huang, Aky{\"u}rek, Anandkumar, Andreas, Mordatch, Torralba, and Zhu]{li2022pretrained}
Shuang Li, Xavier Puig, Chris Paxton, Yilun Du, Clinton Wang, Linxi Fan, Tao Chen, De-An Huang, Ekin Aky{\"u}rek, Anima Anandkumar, Jacob Andreas, Igor Mordatch, Antonio Torralba, and Yuke Zhu.
\newblock Pre-trained language models for interactive decision-making.
\newblock In Alice~H. Oh, Alekh Agarwal, Danielle Belgrave, and Kyunghyun Cho (eds.), \emph{Advances in Neural Information Processing Systems}, 2022{\natexlab{a}}.
\newblock URL \url{https://openreview.net/forum?id=FWMQYjFso-a}.

\bibitem[Li et~al.(2022{\natexlab{b}})Li, Kuncoro, Hoffmann, de~Masson~d{'}Autume, Blunsom, and Nematzadeh]{li-etal-2022-systematic}
Xiang~Lorraine Li, Adhiguna Kuncoro, Jordan Hoffmann, Cyprien de~Masson~d{'}Autume, Phil Blunsom, and Aida Nematzadeh.
\newblock A systematic investigation of commonsense knowledge in large language models.
\newblock In \emph{Proceedings of the 2022 Conference on Empirical Methods in Natural Language Processing}, pp.\  11838--11855, Abu Dhabi, United Arab Emirates, December 2022{\natexlab{b}}. Association for Computational Linguistics.
\newblock \doi{10.18653/v1/2022.emnlp-main.812}.
\newblock URL \url{https://aclanthology.org/2022.emnlp-main.812}.

\bibitem[Liang et~al.(2023)Liang, Huang, Xia, Xu, Hausman, Ichter, Florence, and Zeng]{liang2023code}
Jacky Liang, Wenlong Huang, Fei Xia, Peng Xu, Karol Hausman, Brian Ichter, Pete Florence, and Andy Zeng.
\newblock Code as policies: Language model programs for embodied control, 2023.

\bibitem[Lin et~al.(2023{\natexlab{a}})Lin, Fu, Yang, Ammanabrolu, Brahman, Huang, Bhagavatula, Choi, and Ren]{lin2023swiftsage}
Bill~Yuchen Lin, Yicheng Fu, Karina Yang, Prithviraj Ammanabrolu, Faeze Brahman, Shiyu Huang, Chandra Bhagavatula, Yejin Choi, and Xiang Ren.
\newblock Swiftsage: A generative agent with fast and slow thinking for complex interactive tasks, 2023{\natexlab{a}}.

\bibitem[Lin et~al.(2023{\natexlab{b}})Lin, Huang, Liu, Gu, Sommerer, and Ren]{lin2023grounded}
Bill~Yuchen Lin, Chengsong Huang, Qian Liu, Wenda Gu, Sam Sommerer, and Xiang Ren.
\newblock On grounded planning for embodied tasks with language models.
\newblock In \emph{Proceedings of the AAAI Conference on Artificial Intelligence}, volume~37, pp.\  13192--13200, 2023{\natexlab{b}}.

\bibitem[Logeswaran et~al.(2022)Logeswaran, Fu, Lee, and Lee]{logeswaran-etal-2022-shot}
Lajanugen Logeswaran, Yao Fu, Moontae Lee, and Honglak Lee.
\newblock Few-shot subgoal planning with language models.
\newblock In \emph{Proceedings of the 2022 Conference of the North American Chapter of the Association for Computational Linguistics: Human Language Technologies}, pp.\  5493--5506, Seattle, United States, July 2022. Association for Computational Linguistics.
\newblock \doi{10.18653/v1/2022.naacl-main.402}.
\newblock URL \url{https://aclanthology.org/2022.naacl-main.402}.

\bibitem[Long(2023)]{long2023large}
Jieyi Long.
\newblock Large language model guided tree-of-thought, 2023.

\bibitem[Mu et~al.(2023)Mu, Zhang, Hu, Wang, Ding, Jin, Wang, Dai, Qiao, and Luo]{mu2023embodiedgpt}
Yao Mu, Qinglong Zhang, Mengkang Hu, Wenhai Wang, Mingyu Ding, Jun Jin, Bin Wang, Jifeng Dai, Yu~Qiao, and Ping Luo.
\newblock Embodiedgpt: Vision-language pre-training via embodied chain of thought, 2023.

\bibitem[Pallagani et~al.(2023)Pallagani, Muppasani, Murugesan, Rossi, Srivastava, Horesh, Fabiano, and Loreggia]{pallagani2023understanding}
Vishal Pallagani, Bharath Muppasani, Keerthiram Murugesan, Francesca Rossi, Biplav Srivastava, Lior Horesh, Francesco Fabiano, and Andrea Loreggia.
\newblock Understanding the capabilities of large language models for automated planning.
\newblock \emph{arXiv preprint arXiv:2305.16151}, 2023.

\bibitem[Puig et~al.(2018)Puig, Ra, Boben, Li, Wang, Fidler, and Torralba]{Puig_2018_CVPR}
Xavier Puig, Kevin Ra, Marko Boben, Jiaman Li, Tingwu Wang, Sanja Fidler, and Antonio Torralba.
\newblock Virtualhome: Simulating household activities via programs.
\newblock In \emph{Proceedings of the IEEE Conference on Computer Vision and Pattern Recognition (CVPR)}, June 2018.

\bibitem[Raman et~al.(2022)Raman, Cohen, Rosen, Idrees, Paulius, and Tellex]{DBLP:journals/corr/abs-2211-09935}
Shreyas~Sundara Raman, Vanya Cohen, Eric Rosen, Ifrah Idrees, David Paulius, and Stefanie Tellex.
\newblock Planning with large language models via corrective re-prompting.
\newblock \emph{CoRR}, abs/2211.09935, 2022.
\newblock \doi{10.48550/arXiv.2211.09935}.
\newblock URL \url{https://doi.org/10.48550/arXiv.2211.09935}.

\bibitem[Shinn et~al.(2023)Shinn, Labash, and Gopinath]{shinn2023reflexion}
Noah Shinn, Beck Labash, and Ashwin Gopinath.
\newblock Reflexion: an autonomous agent with dynamic memory and self-reflection.
\newblock \emph{arXiv preprint arXiv:2303.11366}, 2023.

\bibitem[Singh et~al.(2022)Singh, Blukis, Mousavian, Goyal, Xu, Tremblay, Fox, Thomason, and Garg]{singh2022progprompt}
Ishika Singh, Valts Blukis, Arsalan Mousavian, Ankit Goyal, Danfei Xu, Jonathan Tremblay, Dieter Fox, Jesse Thomason, and Animesh Garg.
\newblock Progprompt: Generating situated robot task plans using large language models, 2022.

\bibitem[Song et~al.(2023)Song, Wu, Washington, Sadler, Chao, and Su]{song2023llmplanner}
Chan~Hee Song, Jiaman Wu, Clayton Washington, Brian~M. Sadler, Wei-Lun Chao, and Yu~Su.
\newblock Llm-planner: Few-shot grounded planning for embodied agents with large language models, 2023.

\bibitem[Sun et~al.(2023{\natexlab{a}})Sun, Zhuang, Kong, Dai, and Zhang]{sun2023adaplanner}
Haotian Sun, Yuchen Zhuang, Lingkai Kong, Bo~Dai, and Chao Zhang.
\newblock Adaplanner: Adaptive planning from feedback with language models.
\newblock \emph{arXiv preprint arXiv:2305.16653}, 2023{\natexlab{a}}.

\bibitem[Sun et~al.(2023{\natexlab{b}})Sun, Liu, Wang, Zhu, and Iyyer]{sun2023pearl}
Simeng Sun, Yang Liu, Shuohang Wang, Chenguang Zhu, and Mohit Iyyer.
\newblock Pearl: Prompting large language models to plan and execute actions over long documents.
\newblock \emph{arXiv preprint arXiv:2305.14564}, 2023{\natexlab{b}}.

\bibitem[Suzgun et~al.(2022)Suzgun, Melas-Kyriazi, and Jurafsky]{suzgun2022promptandrerank}
Mirac Suzgun, Luke Melas-Kyriazi, and Dan Jurafsky.
\newblock Prompt-and-rerank: A method for zero-shot and few-shot arbitrary textual style transfer with small language models.
\newblock In \emph{arXiv}, 2022.

\bibitem[Vemprala et~al.(2023)Vemprala, Bonatti, Bucker, and Kapoor]{vemprala2023chatgpt}
Sai Vemprala, Rogerio Bonatti, Arthur Bucker, and Ashish Kapoor.
\newblock Chatgpt for robotics: Design principles and model abilities.
\newblock \emph{Microsoft Autonomous Systems and Robotics Research}, 2023.

\bibitem[Wang et~al.(2020)Wang, Li, Khabsa, Fang, and Ma]{wang2020linformer}
Sinong Wang, Belinda~Z. Li, Madian Khabsa, Han Fang, and Hao Ma.
\newblock Linformer: Self-attention with linear complexity, 2020.

\bibitem[Wang et~al.(2023{\natexlab{a}})Wang, Wei, Schuurmans, Le, Chi, Narang, Chowdhery, and Zhou]{wang2023selfconsistency}
Xuezhi Wang, Jason Wei, Dale Schuurmans, Quoc~V Le, Ed~H. Chi, Sharan Narang, Aakanksha Chowdhery, and Denny Zhou.
\newblock Self-consistency improves chain of thought reasoning in language models.
\newblock In \emph{The Eleventh International Conference on Learning Representations}, 2023{\natexlab{a}}.
\newblock URL \url{https://openreview.net/forum?id=1PL1NIMMrw}.

\bibitem[Wang et~al.(2023{\natexlab{b}})Wang, Cai, Liu, Ma, and Liang]{wang2023describe}
Zihao Wang, Shaofei Cai, Anji Liu, Xiaojian Ma, and Yitao Liang.
\newblock Describe, explain, plan and select: Interactive planning with large language models enables open-world multi-task agents, 2023{\natexlab{b}}.

\bibitem[Wu et~al.(2023{\natexlab{a}})Wu, Min, Bisk, Salakhutdinov, Azaria, Li, Mitchell, and Prabhumoye]{wu2023plan}
Yue Wu, So~Yeon Min, Yonatan Bisk, Ruslan Salakhutdinov, Amos Azaria, Yuanzhi Li, Tom Mitchell, and Shrimai Prabhumoye.
\newblock Plan, eliminate, and track -- language models are good teachers for embodied agents, 2023{\natexlab{a}}.

\bibitem[Wu et~al.(2023{\natexlab{b}})Wu, Wang, Xu, Lu, and Yan]{wu2023embodied}
Zhenyu Wu, Ziwei Wang, Xiuwei Xu, Jiwen Lu, and Haibin Yan.
\newblock Embodied task planning with large language models, 2023{\natexlab{b}}.

\bibitem[Xu et~al.(2019)Xu, Martín-Martín, Huang, Zhu, Savarese, and Fei-Fei]{Xu_Martín-Martín_Huang_Zhu_Savarese_Fei-Fei_2019}
Danfei Xu, Roberto Martín-Martín, De-An Huang, Yuke Zhu, Silvio Savarese, and Li~Fei-Fei.
\newblock Regression planning networks.
\newblock \emph{arXiv: Artificial Intelligence,arXiv: Artificial Intelligence}, Sep 2019.

\bibitem[Yang et~al.(2023)Yang, Nachum, Du, Wei, Abbeel, and Schuurmans]{yang2023foundation}
Sherry Yang, Ofir Nachum, Yilun Du, Jason Wei, Pieter Abbeel, and Dale Schuurmans.
\newblock Foundation models for decision making: Problems, methods, and opportunities, 2023.

\bibitem[Yao et~al.(2022)Yao, Zhao, Yu, Du, Shafran, Narasimhan, and Cao]{yao2022react}
Shunyu Yao, Jeffrey Zhao, Dian Yu, Nan Du, Izhak Shafran, Karthik Narasimhan, and Yuan Cao.
\newblock React: Synergizing reasoning and acting in language models.
\newblock \emph{arXiv preprint arXiv:2210.03629}, 2022.

\bibitem[Yao et~al.(2023)Yao, Yu, Zhao, Shafran, Griffiths, Cao, and Narasimhan]{yao2023tree}
Shunyu Yao, Dian Yu, Jeffrey Zhao, Izhak Shafran, Thomas~L. Griffiths, Yuan Cao, and Karthik Narasimhan.
\newblock Tree of thoughts: Deliberate problem solving with large language models, 2023.

\bibitem[Zeng et~al.(2023)Zeng, Attarian, brian ichter, Choromanski, Wong, Welker, Tombari, Purohit, Ryoo, Sindhwani, Lee, Vanhoucke, and Florence]{zeng2023socratic}
Andy Zeng, Maria Attarian, brian ichter, Krzysztof~Marcin Choromanski, Adrian Wong, Stefan Welker, Federico Tombari, Aveek Purohit, Michael~S Ryoo, Vikas Sindhwani, Johnny Lee, Vincent Vanhoucke, and Pete Florence.
\newblock Socratic models: Composing zero-shot multimodal reasoning with language.
\newblock In \emph{The Eleventh International Conference on Learning Representations}, 2023.
\newblock URL \url{https://openreview.net/forum?id=G2Q2Mh3avow}.

\bibitem[Zhao et~al.(2023)Zhao, Lee, and Hsu]{zhao2023large}
Zirui Zhao, Wee~Sun Lee, and David Hsu.
\newblock Large language models as commonsense knowledge for large-scale task planning, 2023.

\end{thebibliography}
\bibliographystyle{iclr2024_conference}

\newpage
\appendix

{\textbf{\LARGE Appendix}}

\section{Environment}

\subsection{Action Space}
\label{appendix:action-types}
The action types in the Virtual Home Environment are listed as follows:
\begin{multicols}{4}
\begin{enumerate}
    \item Sleep
    \item StandUp
    \item WakeUp
    \item Walk
    \item Find
    \item Grab
    \item Wash
    \item Wipe
    \item Pull
    \item Push
    \item Pour
    \item TurnTo
    \item PointAt
    \item Watch
    \item Touch
    \item Open
    \item Close
    \item Run
    \item Sit
    \item Read
    \item PutOn
    \item Drop
    \item Lie
    \item SwitchOn
    \item SwitchOff
    \item Drink
    \item PutIn
    \item PutBack
\end{enumerate}
\end{multicols}

\begin{figure}[htbp]
    \centering
    \includegraphics[width=0.5\textwidth]{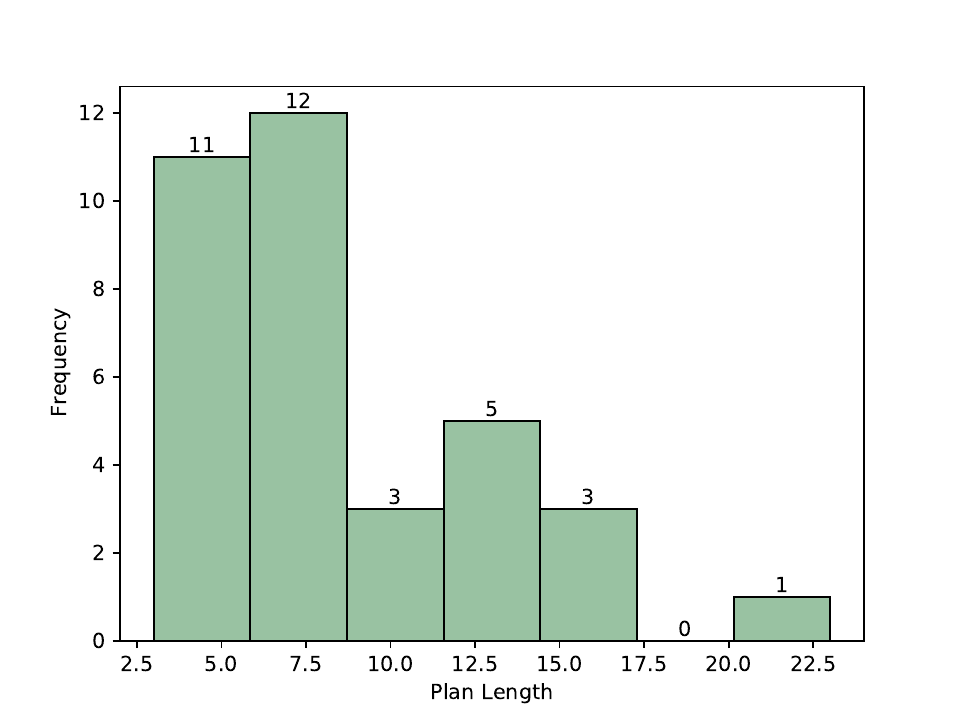}
    \caption{Distribution of plan length.}
    \label{fig:plan_length_hist}
\end{figure}

\subsection{Partial Observation}
We implemented partial observation based on~\citet{li2022pretrained}.
The official definition of partial observation is that when the agent is situated in a room, its observation consists of all the objects in the room that are not located inside enclosed objects. For instance, if an apple is placed inside a closed refrigerator, it will not appear in the observation of the agent.

\subsection{Observation Representation}
In the VirtualHome Environment~\citep{Puig_2018_CVPR}, observations primarily consist of two components: \textit{object states} and \textit{inter-object relationships}.
The \textit{object states} describes the state in which an object exists. For example, the state of television can either be ``on" or ``off", i.e., ``\textit{On(TV)}" or ``\textit{Off(TV)}".
The \textit{inter-object relationships} are represented using predicates to express the relationships between objects. For example, when a character is in close proximity to a television, there may be a predicate such as: ``\textit{Close(character, TV)}".
We convert the observations from VH into English sentences using a rule-based approach. For example, the predicate ``\textit{Close(character, TV)}" is transformed into ``character is close to TV", and the predicate ``\textit{"Off(TV)}" is transformed into "TV is off".

\subsection{Basic Statistics}
\begin{figure}[htbp]
  \begin{minipage}[b]{0.48\textwidth}
    \centering
    \includegraphics[width=0.89\textwidth]{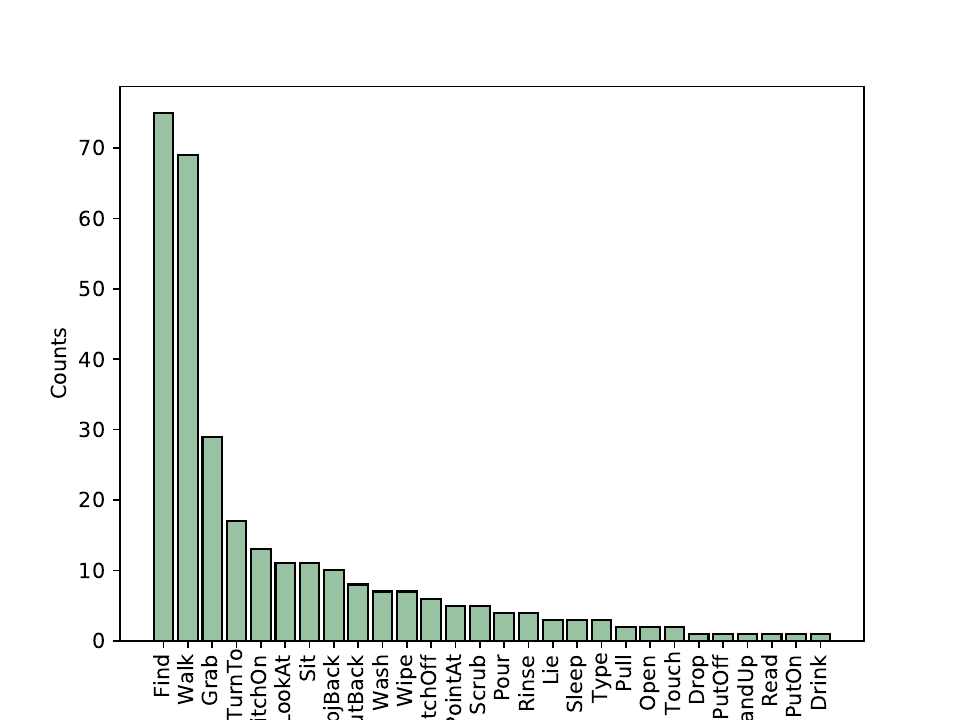}
    \caption{Histogram of action frequencies. Certain actions exhibit a significantly higher frequency than others, such as \textbf{Find} and \textbf{Walk}.}
    \label{fig:action-counts}
  \end{minipage}%
  \hfill
\begin{minipage}[b]{0.48\textwidth}
    \centering
    \includegraphics[width=0.89\textwidth]{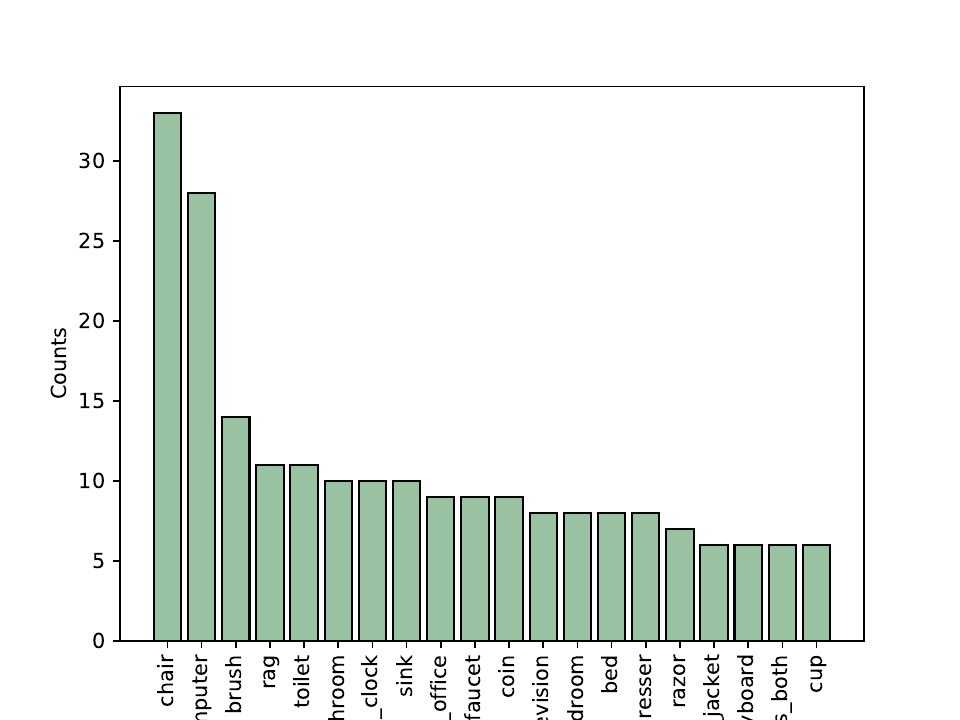}
    \caption{Histogram of object frequencies. Only objects with a frequency in the top 20 are included for better visualization.}
    \label{fig:object-counts}
  \end{minipage}%
\end{figure}

In the gold plan annotated in the dataset, the plan with the longest length consists of 23 actions, while the average length is 8.13. The frequency distribution histogram regarding the length of the plan is shown in Figure~\ref{fig:plan_length_hist}. Furthermore, we have also computed the frequency histograms for actions and objects, which are depicted in Figure~\ref{fig:action-counts} and Figure~\ref{fig:object-counts}.

\section{More Implementation Details}
\label{appendix:implementation-details}
\subsection{Hyper-parameter Search}
For both Grounded Deciding and \sbs, we conducted a grid search over the sampling parameters of OpenAI APIs. 
The search range for temperature was set from 0.1 to 1.0 in increments of 0.1, while the search range for top-p was set to 0.85, 0.9, 0.95, and 1.0.

In the case of Grounded Deciding, the optimal hyperparameter combination was found to be a temperature of 0.7 and topp of 1.0. As for \sbs, the optimal hyperparameter combination was a temperature of 0 and topp of 1.0.

\subsection{Baseline Models}
\label{appendix:baseline-models-details}
\noindent \textbf{\zeroshot}
~\citep{DBLP:conf/icml/HuangAPM22} propose to translate each unstructured action generated by LLM into an admissible action via another pre-trained masked language model. The translated action is then appended to the prompt used for generating the remaining steps. We utilize the official implementation provided by \cite{DBLP:conf/icml/HuangAPM22}, which employs a dynamically retrieved plan as an exemplar in the prompt. Moreover, concerning the hyper-parameters, we configure the maximum number of steps as 20 and set the early stopping threshold to 0.5 to achieve optimal performance.

\noindent \textbf{\progprompt}
~\citep{singh2022progprompt} proposes a programming language-inspired prompt with an assert statement. These assert statements provide a mechanism for incorporating environment feedback into the plan generation process, ensuring that preconditions for each action are met;

\subsection{Post-Processing}
To ensure the generated plan is actionable in the environment, we further implement a post-processing module after plan sampling: (i) \textbf{Format Checking:} Actions that do not conform to the required format and thus cannot be parsed, is discarded. Given the strong format following ability of GPT-3.5, the number of discarded items is minimal; (ii) \textbf{Object \& Action Translation:} Even with the correct format, the plan generated by the LLM might include actions or objects not present in the environment. This issue often arises from semantically accurate but not exactly matching results. For instance, if the environment's object name is \textit{"food\_egg"}, but the generated action includes \textit{"egg"}, this discrepancy requires resolution. Firstly, we parse the LLM's action string to identify the action and object names. Then we use BERT similarity to match these with the environment's available actions and objects~\footnote{\url{https://www.sbert.net/}}. For example, for the LLM's generated string \textit{'[Switch On] \textless telly \textgreater'}, the action parser identifies \textit{"Switch On"} as the action and \textit{"telly"} as the object. These are then matched with available actions and objects to result in the actionable action \textit{"[SwitchOn] \textless TV \textgreater"}.

\section{More Experimental Results}

\subsection{Results by Plan Length}

\begin{table}[htbp]
  \centering
    \begin{tabular}{lcccc}
    \toprule
          & \exec $\uparrow$  & \sr $\uparrow$   & \gcr $\uparrow$  & \nocorrection $\downarrow$ \\
    \midrule
    $0 < |a| \leq 5$     & 100.00 $\pm$ 0.00   & 64.72$\pm$5.20 & 77.12$\pm$4.13 & 0.30$\pm$0.15 \\
    $5 < |a| \leq 10$      & 83.59$\pm$4.03 & 35.10$\pm$5.95 & 52.25$\pm$3.33 & 2.47$\pm$0.35 \\
    $10 < |a| \leq 15$      & 88.09$\pm$8.48 & 26.98$\pm$4.89 & 55.98$\pm$7.00 & 3.20$\pm$1.12 \\
    $15 < |a|$      & 66.67$\pm$0.00 & 22.22$\pm$0.00 & 36.11$\pm$0.00 & 4.09$\pm$0.21 \\
    \bottomrule
    \end{tabular}%
    \caption{The performance of $\ours_{N=50}$ across different plan sequence lengths.}
  \label{tab:plan-length}%
\end{table}%

Table~\ref{tab:plan-length} above presents the performance of $\ours_{N=50}$, categorized by different plan sequence lengths. 
In general, as the plan sequence length increases, the performance of the model tends to decrease.
Specifically, for tasks with plan lengths smaller than 5, the \sr can reach 64.72\% and even 100\% for \exec. However, for tasks with plan lengths larger than 15, the \sr decreases to 22.22\% (-42.5\%), and \exec decreases to 66.67\% (-33.33\%).
Furthermore, as the plan length increases, the number of corrections in the model also increases. This is due to the higher likelihood of errors accumulating in longer sequential task planning, thus necessitating a greater need for error correction.
The above experimental results provide a more comprehensive analysis of the performance of our approach, emphasizing potential directions for future enhancements.
Furthermore, we have provided the results of \localreplan as shown in Table~\ref{tab:plan-length-sbs}. It can be observed that while $\ours$ maintains comparable performance across different plan lengths, it exhibits a significant advantage in terms of the number of corrections.

\begin{table}[htbp]
  \centering
    \begin{tabular}{lcccc}
    \toprule
          & \exec $\uparrow$  & \sr $\uparrow$   & \gcr $\uparrow$  & \nocorrection $\downarrow$ \\
    \midrule
    $0 < |a| \leq 5$     & 94.17 $\pm$ 2.04   & 56.94 $\pm$ 0.79 & 69.46 $\pm$ 1.70 & 1.05 $\pm$ 0.32 \\
    $5 < |a| \leq 10$    & 77.02 $\pm$ 7.63 & 38.10 $\pm$ 3.57 & 49.66 $\pm$ 5.45 & 3.92 $\pm$ 0.56 \\
    $10 < |a| \leq 15$   & 69.84 $\pm$ 2.97 & 24.78 $\pm$ 5.83 & 44.52 $\pm$ 2.16 & 5.14 $\pm$ 0.17 \\
    $15 < |a|$           & 72.22 $\pm$ 28.33 & 22.22 $\pm$ 0.00 & 35.65 $\pm$ 8.36 & 4.83 $\pm$ 0.87 \\
    \bottomrule
    \end{tabular}%
    \caption{The performance of $\localreplan$ across different plan sequence lengths.}
  \label{tab:plan-length-sbs}%
\end{table}%

\subsection{Results by Scene}

As is shown in Table~\ref{tab:diff-scene}, \ours exhibits consistent performance across various scenes, thereby further illustrating its robustness.

\begin{table}[htbp]
  \centering
    \begin{tabular}{lcccc}
    \toprule
          & \exec $\uparrow$  & \sr $\uparrow$    & \gcr $\uparrow$  & \nocorrection $\downarrow$ \\
    \midrule
    ENV-1 & 92.42$\pm$2.14 & 45.45$\pm$3.71 & 64.72$\pm$3.25 & 1.74$\pm$0.44 \\
    ENV-2 & 91.30$\pm$4.10 & 36.75$\pm$4.10 & 52.43$\pm$7.13 & 2.33$\pm$0.47 \\
    ENV-3 & 85.18$\pm$2.62 & 37.04$\pm$2.62 & 50.80$\pm$3.19 & 1.83$\pm$0.39 \\
    ENV-4 & 88.89$\pm$3.14 & 48.89$\pm$3.14 & 66.74$\pm$1.72 & 2.35$\pm$0.58 \\
    \bottomrule
    \end{tabular}%
  \caption{The performance of $\ours_{N=50}$ across different scenes.}
  \label{tab:diff-scene}%
\end{table}%

\subsection{Diversity of Sampled Plans}
\begin{figure}[htbp]
    \centering
    \includegraphics[width=0.5\textwidth]{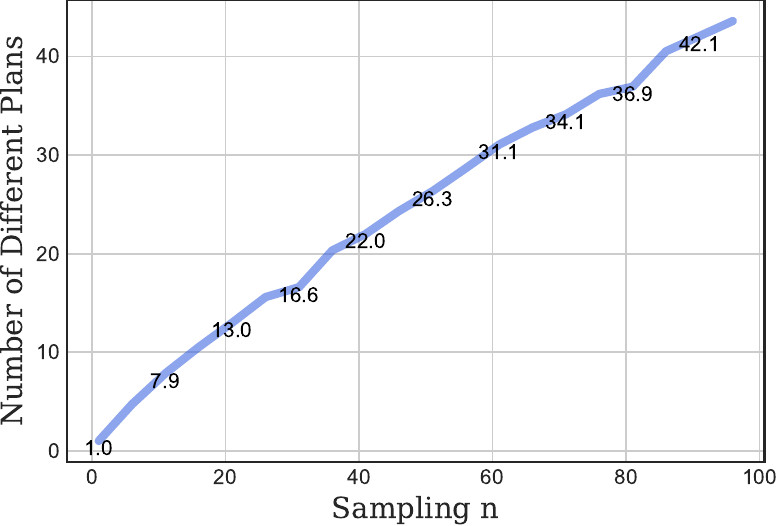}
    \caption{The diversity of plans generated. The x-axis represents the number of programs sampled, while the y-axis represents the average number of different plans generated.}
    \label{fig:diversity-plan}
\end{figure}

As shown in Figure~\ref{fig:diversity-plan}, the number of different plans varies approximately linearly with the change in the sampling n.
Therefore, when conducting plan sampling, the issue of homogenization due to the sampled plans will not arise.

\subsection{Upper Limit of Plan Sampling}

\begin{figure}[htbp]
    \centering
    \includegraphics[width=0.5\textwidth]{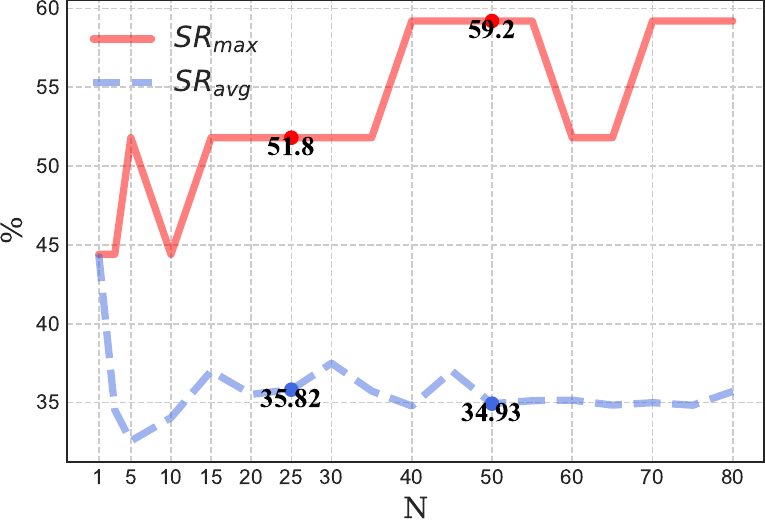}
    \caption{Maximum and average Success Rate (SR) for all sampled plans.}
    \label{fig:upper-bound-ps-sr}
\end{figure}

We also compute the corresponding Success Rate (SR) result of Figure~\ref{fig:upper-bound-ps} by: 
\textbf{(i)} the maximum SR for all generated plans, i.e., {\small $SR_{max}(\boldsymbol{c}) = \max\limits_{i=1}^{N} (SR(c_i))$};
\textbf{(ii)} the average SR for all generated plans, i.e., {\small $SR_{avg}(\boldsymbol{c}) = \frac{1}{N} \sum_{i=1}^{N} (SR(c_i))$}.
As is shown in Figure~\ref{fig:upper-bound-ps-sr}, the trend of SR concerning N closely aligns with the trajectory of GCR.

\subsection{Experimental Results with Best-First Search}

\noindent \textbf{Language-Guided Best-First Search}
The edges of the action tree are assigned weights based on the log probabilities of the actions they lead to. These weights are aggregated when multiple paths converge into the same action, thereby reflecting the cumulative likelihood of reaching that action within the context of the tree's structure. The evaluation function of Language-Guided Best-First Search is defined as: $f_{\text{L}}(a_i) = \frac{e^{\log(\text{Prob}(a_i))}}{\sum_{j} e^{\log(\text{Prob}(a_j))}}$. Here, $a_i$ represents one of the child nodes of a given node and iterates over all child nodes. The softmax function normalizes the log probabilities, making them suitable for probabilistic comparison and selection.

\noindent \textbf{Environment-Guided Best-First Search}
Actions are weighted based on the observability of their associated objects. If all objects involved in an action are observable in the environment, the action is assigned a higher weight (1), otherwise, a lower weight (0). To facilitate comparative analysis and to account for varying numbers of actions and objects, a softmax function is applied to the weights of actions emanating from each node. The evaluation function of Environment-Guided Best-First Search is defined as $f_{\text{E}}(a_i) = \frac{e^{\text{Observability}(a_i)}}{\sum_{j} e^{\text{Observability}(a_j)}}$. 

\noindent \textbf{Hybrid Best-First Search}
The Hybrid Method combines the probabilistic language and environment observability factors. The evaluation function is:
$f_{\text{Hybrid}}(a_i) = \alpha f_{\text{L}}(a_i) + (1-\alpha)  f_{\text{E}}(a_i) $
In this function, $\alpha$ is a hyper-parameter between 0 and 1 that balances the influence of the language-based probabilities and the environmental observability

\noindent \textbf{Experimental Setup}
We conducted experiments in the case of N=50 without error correction. As for the hyper-parameter $\alpha$, we heuristically set it to 0.5.

\noindent \textbf{Experimental Results}
As demonstrated by Table~\ref{tab:results-bfs}, although heuristic methods exhibit a greater advantage in terms of token efficiency, their SR falls short compared to grounded deciding (using LLM as an implicit evaluation function). Moreover, combining both evaluation functions yields even greater performance improvements, thus demonstrating that there is still room for enhancement through the optimization of the evaluation function in heuristic methods.

\begin{table*}[htbp]
  \centering
    \caption{Experimental Results with Best-First Search}
    \begin{tabular}{lcccc}
    \toprule
          & \exec $\uparrow$ & \sr $\uparrow$ & \gcr $\uparrow$ & \cost $\downarrow$ \\
    \midrule
    $\ours$   & 49.01$\pm$5.67 & 28.14$\pm$2.45 & 35.84$\pm$4.20 & 3.48$\pm$0.04  \\
    $\textsc{LanguageGuided}$ & 34.71$\pm$0.65 & 15.06$\pm$1.02 & 22.47$\pm$1.72 & 2.16$\pm$0.02  \\
    $\textsc{EnvironmentGuided}$ & 37.26$\pm$0.17 & 6.24$\pm$0.65 & 12.10$\pm$1.41 & 2.16$\pm$0.02  \\
    $\textsc{Hybrid}_{\alpha=0.5}$ & 38.71$\pm$0.42 & 17.74$\pm$0.69 & 25.34$\pm$0.95 & 2.16$\pm$0.02  \\
    \bottomrule
    \end{tabular}%
  \label{tab:results-bfs}
\end{table*}


\subsection{Proportion of New Objects}

\begin{figure}[htbp]
    \centering
    \includegraphics[width=1\textwidth]{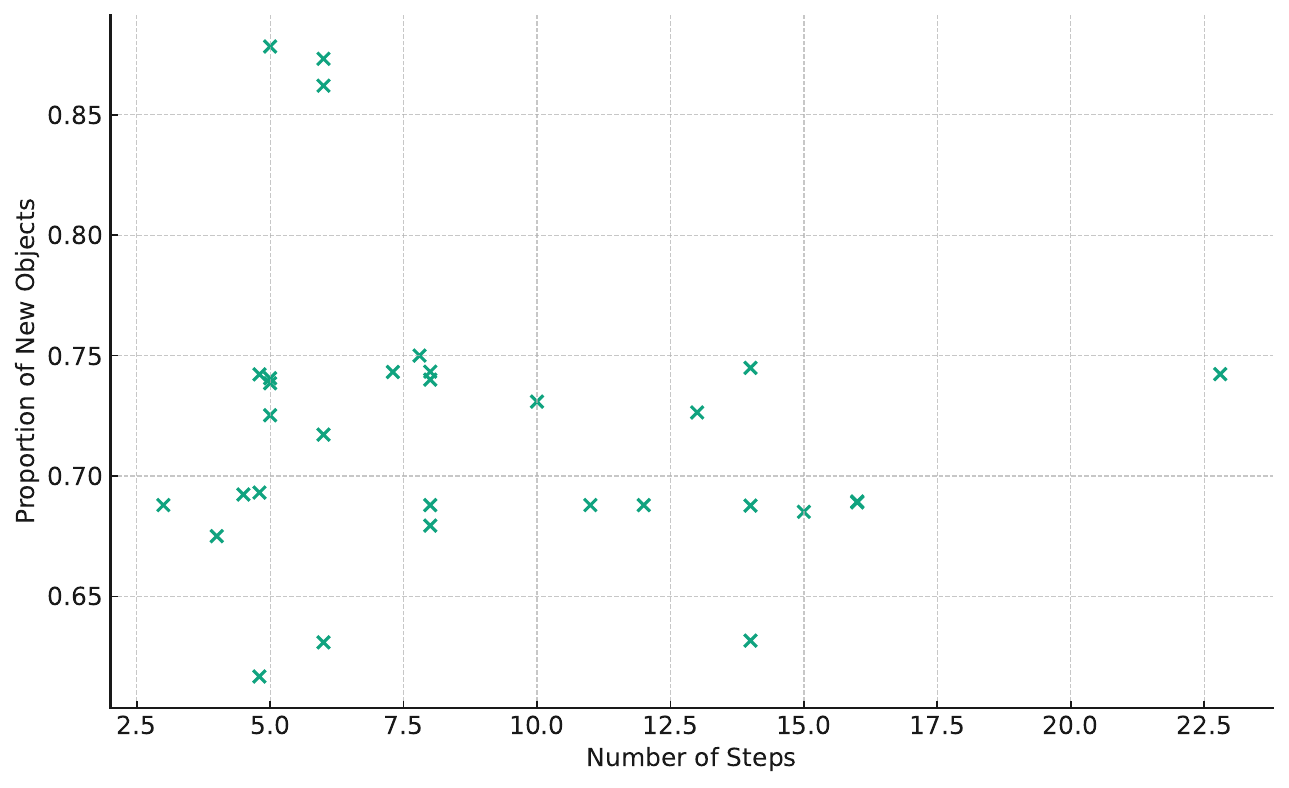}
    \caption{Each point represents a distinct task. The proportion of new objects is calculated with gold plans in the dataset.}
    \label{fig:proportion}
\end{figure}

The relationship between the number of steps and the proportion of new objects in evaluated tasks is shown in Figure~\ref{fig:proportion}.

\section{Details on Token Efficiency}
\label{appendix:token-efficiency}
The detailed derivation of the boundary conditions for $N$ is as follows:
\begin{align*}
& \text{\textit{to prove}} \quad \mu_{ours} < \mu_{sbs} \\
& \Rightarrow \quad \rho_{ps} + MN|a| + M \cdot (\rho_{gd} + N)  < M \cdot (\rho_{ps} + \rho_{gd} + |a|) \\
& \Rightarrow \quad MN|a| + M \cdot N  < (M - 1) \cdot \rho_{ps} + M \cdot |a| \\
& \Rightarrow \quad M \cdot (|a| + 1) \cdot N  < (M - 1) \cdot \rho_{ps} + M \cdot |a| \\
& \Rightarrow \quad N < \frac{1-1/M}{1+1/|a|} \cdot \frac{\rho_{ps}}{|a|} + \frac{|a|}{|a| + 1} \\
\end{align*}

\section{Qualitative Analysis on Errors}
\label{appendix:error-types}

\subsection{Incorrect Deciding}
\noindent \textbf{Task:} 
\textit{Hang up jacket}

\begin{figure}[ht]
    \centering
    \includegraphics[width=1\textwidth]{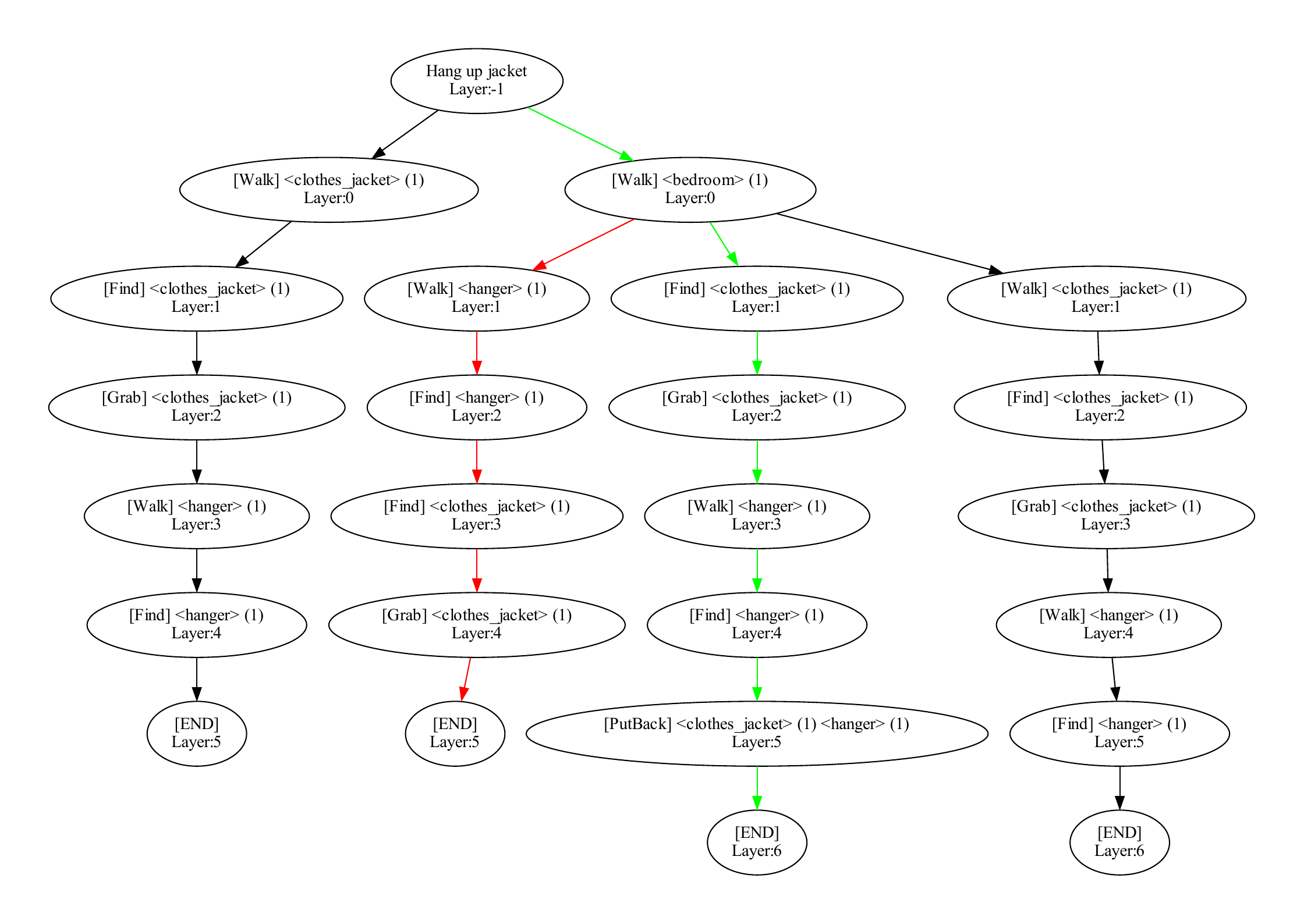}
\caption{Visualization of the action tree for \textit{Hang up jacket}. The \textcolor{red}{red} path is the plan chosen by LLM, while the \textcolor{green}{green} path is the correct plan.}
\label{fig:deciding-error-fig}
\end{figure}

\noindent \textbf{Action Tree:}
See Figure~\ref{fig:deciding-error-fig}

\noindent \textbf{Explanation:}
As depicted in Figure~\ref{fig:deciding-error-fig}, the model made an incorrect decision during the second step. This is due to the failure of \ours to comprehend the sequential order of searching for a jacket and searching for a hanger.

\subsection{Environment Misunderstanding}
\noindent \textbf{Task:} 
\textit{Clean toilet}

\noindent \textbf{Example Plan:}
\begin{multicols}{2}
\begin{enumerate}
\item \textbf{[Walk]} \textless bathroom\textgreater (1)
\item \textbf{[Walk]} \textless toilet\textgreater (1)
\item \textbf{[Pull]} \textless toilet\textgreater (1)
\item \textbf{[Wash]} \textless toilet\textgreater (1)
\item \textbf{[Wipe]} \textless toilet\textgreater (1)
\item \textbf{[Push]} \textless toilet\textgreater (1)
\item \textbf{[Wash]} \textless toilet\textgreater (1)
\item \textbf{[Find]} \textless mop\_bucket\textgreater (1)
\item \textbf{[Walk]} \textless detergent\textgreater (1)
\item \textbf{[Grab]} \textless detergent\textgreater (1)
\item \textbf{[Pour]} \textless detergent\textgreater (1) \\ \textless mop\_bucket\textgreater (1)
\end{enumerate}
\end{multicols}

\noindent \textbf{Explanation:}
The toilet is not movable, so the execution of step 3 would be failed.

\subsection{Incomplete Plan}
\noindent \textbf{Task:} 
\textit{Put alarm clock in bedroom}

\noindent \textbf{Example Plan:}
\begin{enumerate}
\item \textbf{[Walk]} \textless bedroom\textgreater (1)
\item \textbf{[Walk]} \textless alarm\_clock\textgreater (1)
\item \textbf{[Find]} \textless alarm\_clock\textgreater (1)
\item \textbf{[Grab]} \textless alarm\_clock\textgreater (1)
\item \textbf{[Find]} \textless dresser\textgreater (1)
\item \textbf{[Open]} \textless dresser\textgreater (1)
\item \textbf{[SwitchOn]} \textless alarm\_clock\textgreater (1)
\end{enumerate}

\noindent \textbf{Explanation:}
After executing this plan, the alarm clock will be in the hand of the character. However, a step is missing: \textbf{[PutBack]} the \textless alarm\_clock\textgreater (1) on the \textless dresser\textgreater (1) to place the alarm clock \textit{in bedroom}.

\subsection{Illogical Error}
\noindent \textbf{Task:} 
\textit{Shave}

\noindent \textbf{Example Plan:}
\begin{multicols}{2}
\begin{enumerate}
\item \textbf{[Walk]} \textless bathroom\textgreater (1)
\item \textbf{[Find]} \textless shaving\_cream\textgreater (1)
\item \textbf{[Grab]} \textless shaving\_cream\textgreater (1)
\item \textbf{[Walk]} \textless after\_shave\textgreater (1)
\item \textbf{[Find]} \textless after\_shave\textgreater (1)
\item \textbf{[Grab]} \textless after\_shave\textgreater (1)
\item \textbf{[Walk]} \textless razor\textgreater (1)
\item \textbf{[Find]} \textless razor\textgreater (1)
\item \textbf{[Grab]} \textless razor\textgreater (1)
\item \textbf{[PutOn]} \textless shaving\_cream\textgreater (1)
\item \textbf{[Wash]} \textless headset\textgreater (1)
\item \textbf{[PutOn]} \textless after\_shave\textgreater (1)
\end{enumerate}
\end{multicols}

\noindent \textbf{Explanation:}
The character does not have a free hand when executing step 9. This is because, in steps 3 and 6, the character has already picked up two items.

\subsection{Semantically Correct}
\noindent \textbf{Task:} 
\textit{Use computer}

\noindent \textbf{Example Plan:}
\begin{enumerate}
\item \textbf{[Walk]} \textless home\_office\textgreater (1)
\item \textbf{[Walk]} \textless computer\textgreater (1)
\item \textbf{[Find]} \textless chair\textgreater (1)
\item \textbf{[Sit]} \textless chair\textgreater (1)
\item \textbf{[SwitchOn]} \textless computer\textgreater (1)
\item \textbf{[Type]} \textless keyboard\textgreater (1)
\end{enumerate}


\noindent \textbf{Explanation:}
The error information from the environment is as follows: ``\textless keyboard \textgreater" cannot be found when executing ``\textbf{[Type]} \textless keyboard \textgreater (1)". However, at this moment, the character is seated in front of the computer, indicating that the keyboard should be located near the character.

\subsection{More Results on Error Types}

As illustrated in Table~\ref{tab:error-types-with-correction}, we conducted a further analysis of the error types in the $\ours_{N=25}$ model \textit{with correction}. The distribution is strikingly similar to that of the model \textit{without correction}. Moreover, owing to the presence of error correction, the model demonstrates enhanced capability in avoiding \textit{semantically correct} errors during the grounded deciding stage.

\begin{table}[b]
  \centering
  \small
  \caption{Distribution of error types of the $\ours_{N=25}$ \textit{with correction} model.}
  \vspace{-6pt}
    \begin{tabular}{lp{6cm}l}
    \toprule
    Error Type & Explanation & \multicolumn{1}{l}{Proportion(\%)} \\
    \midrule
    Missing Correct Plans & Plan sampling did not yield correct plans & \textbf{52.6\%} \\
    ~~~~Environment Misunderstanding & Misunderstandings on actions or objects & ~~~~21.1\% \\
    ~~~~Incomplete Plan & The absence of essential steps & ~~~~15.8\% \\
    ~~~~Illogical Error & The generated plan is logically incorrect & ~~~~5.3\% \\
    ~~~~Semantically Correct & Execution failed but semantically correct & ~~~~10.5\% \\
    Grounded Deciding Error & Errors during grounded deciding & \textbf{47.4\%} \\
    ~~~~Incorrect Deciding & Incorrect decisions at specific nodes & ~~~~47.4\% \\
    ~~~~Semantically Correct & Execution failed but semantically correct & ~~~~0.0\% \\
    \bottomrule
    \end{tabular}%
  \label{tab:error-types-with-correction}%
\end{table}%

\section{Prompts}
\label{appendix:prompt-examples}

\subsection{\sbs}

\textbf{\textit{(Instruction)}}
\textcolor[rgb]{0.545,0,0}{
You need to act as a task planner who decomposes a high-level household task into several sub-tasks. The temporal relationship between subtask sequences must adhere to common-sense logic.
Each sub-task can be one of the following form: 1. [action\_name]; 2. [action\_name] \textless object name 1\textgreater (object id 1). 3. [action\_name] \textless object name 1\textgreater (object id 1) \textless object name 2\textgreater (object id 2). The number of arguments depends on the action type.
The (object id) is used to tell the simulator that the actions should be done on the same object instance. For example a program as:
[Walk] \textless glass\textgreater (1)
[Grab] \textless glass\textgreater (1)
Indicates that the agent should first walk to a glass and then grab that same glass.
If you think your task has been successful, you can output [END], which is action type 1.
}

\textbf{\textit{(Global Information)}}
\textcolor[rgb]{0,0,0.545}{
For action type 1, the available actions are: [Sleep], [StandUp], [WakeUp]
For action type 2, the available actions are: [Walk], [Find], [Grab], [Wash], [Wipe], [Pull], [Push], [Pour], [TurnTo], [PointAt], [Watch], [Touch], [Open], [Close], [Run], [Sit], [Read], [PutOn], [Drop], [Lie], [SwitchOn], [SwitchOff], [Drink]
For action type 3, the available actions are: [PutIn], [PutBack]
All action\_name of the sub-tasks must be chosen from the above actions, and follow the corresponding format.
You are in a house that consists of four rooms. These rooms are bathroom, dining\_room, bedroom, home\_office.
Available objects in the house are : clothes\_hat, ceilinglamp, cpuscreen, orchid, couch, trashcan, dresser, dishwasher, centerpiece, phone, toaster, measuring\_cup, stereo, mat, computer, envelope, oven\_mitts, piano\_bench, box, photoframe, shower, ceiling, wall, window, freezer, faucet, detergent, light, desk, napkin, food\_rice, kitchen\_counter, folder, stovefan, walllamp, food\_food, coffee\_pot, food\_steak, jelly, vacuum\_cleaner, powersocket, filing\_cabinet, alcohol, bathroom, door, bathroom\_counter, clothes\_gloves, microwave, oven, sink, milk, ice, bedroom, laptop, doorjamb, food\_cake, bills, tea\_bag, television, laser\_pointer, toilet, board\_game, sponge, food\_carrot, table, tray, cupboard, mousepad, picture, tvstand, tablelamp, hanger, pot, dry\_pasta, floor, knifeblock, curtain, chair, food\_bread, drawing, creditcard, check, coffe\_maker, character, pasta, bag, food\_bacon, bookshelf, toothbrush\_holder, cutting\_board, home\_office, dining\_room, nail\_polish, pillow, tape, nightstand, bathroom\_cabinet, bench, conditioner, cat, bed, keyboard, mouse
All object names must be chosen from the above object list
}

\textbf{\textit{(Observation)}}
\textcolor[rgb]{0,0.392,0}{
Currently, you are standing in the bedroom, and holding nothing in your right hand and nothing in your left hand.
 pillow is clean. napkin is clean. pillow is dirty. bed is clean. mat is dirty. pillow is close to drawing. tablelamp is close to bed. mat is facing drawing. pillow is inside bedroom. mat is close to table. bed is close to drawing. table is close to mat. pillow is on floor. floor is close to bed. tablelamp is close to pillow. mat is close to curtain. bed is facing computer. mat is close to floor. pillow is facing drawing. curtain is close to mat. bed is close to tablelamp. wall is close to mat. napkin is inside bedroom. window is close to mat. pillow is close to tablelamp. bed is close to floor. pillow is close to wall. mat is close to filing\_cabinet. drawing is close to bed. pillow is close to floor. bed is close to wall. filing\_cabinet is close to mat. bed is close to nightstand. mat is inside bedroom. pillow is close to pillow. pillow is close to nightstand. mat is close to wall. wall is close to pillow. nightstand is close to pillow. nightstand is close to bed. drawing is close to pillow. floor is close to pillow. bed is inside bedroom. floor is close to mat. wall is close to bed. mat is close to window. pillow,napkin,pillow,mat,bed,pillow,pillow is inside bedroom.}

\textbf{\textit{(In-Context Examples)}} \newline
\textcolor[rgb]{0.412,0.412,0.412}{
Task: Watch TV  \newline
[\textbf{Find}] \textless remote\_control\textgreater (1)\newline
[\textbf{Find}] \textless television\textgreater (1)\newline
[\textbf{SwitchOn}] \textless television\textgreater (1)\newline
[\textbf{Find}] \textless couch\textgreater (1)\newline
[\textbf{Sit}] \textless couch\textgreater (1)\newline
[\textbf{Touch}] \textless remote\_control\textgreater (1)\newline
[\textbf{TurnTo}] \textless television\textgreater (1)\newline
[\textbf{LookAt}] \textless television\textgreater (1) \newline
\newline
Task: Turn on light\newline
[\textbf{Walk}] \textless dining\_room\textgreater (1)\newline
[\textbf{Walk}] \textless light\textgreater (1)\newline
[\textbf{Find}] \textless light\textgreater (1)\newline
[\textbf{SwitchOn}] \textless light\textgreater (1)\newline
[\textbf{Find}] \textless light\textgreater (2)\newline
[\textbf{SwitchOn}] \textless light\textgreater (2) \newline
\newline
Task: Go to sleep\newline
[\textbf{Walk}] \textless bedroom\textgreater (1)\newline
[\textbf{Walk}] \textless bed\textgreater (1)\newline
[\textbf{Lie}] \textless bed\textgreater (1)\newline
[\textbf{Sleep}] \newline
\newline
Task: Brush teeth\newline
[\textbf{Walk}] \textless bathroom\textgreater (1)\newline
[\textbf{Walk}] \textless toothbrush\_holder\textgreater (1)\newline
[\textbf{Find}] \textless toothbrush\_holder\textgreater (1)\newline
[\textbf{Find}] \textless toothbrush\textgreater (1)\newline
[\textbf{Grab}] \textless toothbrush\textgreater (1)\newline
[\textbf{Walk}] \textless tooth\_paste\textgreater (1)\newline
[\textbf{Find}] \textless tooth\_paste\textgreater (1)\newline
[\textbf{Grab}] \textless tooth\_paste\textgreater (1)\newline
[\textbf{Pour}] \textless tooth\_paste\textgreater (1) \textless toothbrush\textgreater (1)\newline
[\textbf{Find}] \textless teeth\textgreater (1)\newline
[\textbf{Scrub}] \textless teeth\textgreater (1) \newline
}

Task: Take nap\newline
[\textbf{Walk}] \textless bedroom\textgreater (1)\newline

\subsection{Plan Sampling}
\textbf{\textit{(Instruction)}}
\textcolor[rgb]{0.545,0,0}{
You need to act as a task planner, who decompose a high-level household task into several sub-tasks. The temporal relationship between subtask sequences must adhere to common-sense logic.
Each sub-task can be one of the following form: 1. [action\_name]; 2. [action\_name] \textless object name 1 \textgreater (object id 1). 3. [action\_name] \textless object name 1 \textgreater (object id 1) \textless object name 2 \textgreater (object id 2). The number of arguments depends on the action type.
The (object id) is used to tell the simulator that the actions should be done on the same object instance. For example a program as:
[Walk] \textless glass\textgreater (1)
[Grab] \textless glass\textgreater (1)
Indicates that the agent should first walk to a glass, and then grab that same glass.
}

\textbf{\textit{(Global Information)}}
\textcolor[rgb]{0,0,0.545}{
For action type 1, the available actions are: [Sleep], [StandUp], [WakeUp]
For action type 2, the available actions are: [Walk], [Find], [Grab], [Wash], [Wipe], [Pull], [Push], [Pour], [TurnTo], [PointAt], [Watch], [Touch], [Open], [Close], [Run], [Sit], [Read], [PutOn], [Drop], [Lie], [SwitchOn], [SwitchOff], [Drink]
For action type 3, the available actions are: [PutIn], [PutBack]
All action\_name of the sub-tasks must be chosen from the above actions, and follow the corresponding format.
You are in a house that consists of four rooms. These rooms are bathroom, dining\_room, bedroom, home\_office.
Available objects in the house are : clothes\_hat, ceilinglamp, cpuscreen, orchid, couch, trashcan, dresser, dishwasher, centerpiece, phone, toaster, measuring\_cup, stereo, mat, computer, envelope, oven\_mitts, piano\_bench, box, photoframe, shower, ceiling, wall, window, freezer, faucet, detergent, light, desk, napkin, food\_rice, kitchen\_counter, folder, stovefan, walllamp, food\_food, coffee\_pot, food\_steak, jelly, vacuum\_cleaner, powersocket, filing\_cabinet, alcohol, bathroom, door, bathroom\_counter, clothes\_gloves, microwave, oven, sink, milk, ice, bedroom, laptop, doorjamb, food\_cake, bills, tea\_bag, television, laser\_pointer, toilet, board\_game, sponge, food\_carrot, table, tray, cupboard, mousepad, picture, tvstand, tablelamp, hanger, pot, dry\_pasta, floor, knifeblock, curtain, chair, food\_bread, drawing, creditcard, check, coffe\_maker, character, pasta, bag, food\_bacon, bookshelf, toothbrush\_holder, cutting\_board, home\_office, dining\_room, nail\_polish, pillow, tape, nightstand, bathroom\_cabinet, bench, conditioner, cat, bed, keyboard, mouse
All object names must be chosen from the above object list
}

\textbf{\textit{(Initial Observation)}}
\textcolor[rgb]{0,0.392,0}{
Currently, you are standing in the home\_office, and holding nothing in your right hand and nothing in your left hand.
}

\textbf{\textit{(In-Context Examples)}} \newline
\textcolor[rgb]{0.412,0.412,0.412}{
Task: Watch TV  \newline
[\textbf{Find}] \textless remote\_control\textgreater (1)\newline
[\textbf{Find}] \textless television\textgreater (1)\newline
[\textbf{SwitchOn}] \textless television\textgreater (1)\newline
[\textbf{Find}] \textless couch\textgreater (1)\newline
[\textbf{Sit}] \textless couch\textgreater (1)\newline
[\textbf{Touch}] \textless remote\_control\textgreater (1)\newline
[\textbf{TurnTo}] \textless television\textgreater (1)\newline
[\textbf{LookAt}] \textless television\textgreater (1) \newline
\newline
Task: Turn on light\newline
[\textbf{Walk}] \textless dining\_room\textgreater (1)\newline
[\textbf{Walk}] \textless light\textgreater (1)\newline
[\textbf{Find}] \textless light\textgreater (1)\newline
[\textbf{SwitchOn}] \textless light\textgreater (1)\newline
[\textbf{Find}] \textless light\textgreater (2)\newline
[\textbf{SwitchOn}] \textless light\textgreater (2) \newline
\newline
Task: Go to sleep\newline
[\textbf{Walk}] \textless bedroom\textgreater (1)\newline
[\textbf{Walk}] \textless bed\textgreater (1)\newline
[\textbf{Lie}] \textless bed\textgreater (1)\newline
[\textbf{Sleep}] \newline
\newline
Task: Brush teeth\newline
[\textbf{Walk}] \textless bathroom\textgreater (1)\newline
[\textbf{Walk}] \textless toothbrush\_holder\textgreater (1)\newline
[\textbf{Find}] \textless toothbrush\_holder\textgreater (1)\newline
[\textbf{Find}] \textless toothbrush\textgreater (1)\newline
[\textbf{Grab}] \textless toothbrush\textgreater (1)\newline
[\textbf{Walk}] \textless tooth\_paste\textgreater (1)\newline
[\textbf{Find}] \textless tooth\_paste\textgreater (1)\newline
[\textbf{Grab}] \textless tooth\_paste\textgreater (1)\newline
[\textbf{Pour}] \textless tooth\_paste\textgreater (1) \textless toothbrush\textgreater (1)\newline
[\textbf{Find}] \textless teeth\textgreater (1)\newline
[\textbf{Scrub}] \textless teeth\textgreater (1) \newline
\newline
}
Task: Take nap

\subsection{Grounded Deciding}
\textbf{\textit{(Instruction)}}
\textcolor[rgb]{0.545,0,0}{
You need to act as a home robot. At each moment, I will provide you with observations of your current environment, as well as the high-level task I want you to do, and previous mid-level sub-tasks that have been executed.
Then, you need to select the best sub-task from the options I provide to complete the designated home task based on the observation and your past experience.
When one choosed sub-task causes an error in the environment, you will be provided with the error information and the corresponding sub-task, and you need to re-choose a corrective sub-task at the current time step. 
For example, The sub-tasks that have been executed in the environment are: \newline
[GRAB] \textless plate\textgreater (1) \newline
[WALK] \textless dining room\textgreater (1) \newline
The choosed sub-task is: 
[PUTBACK] \textless plate\textgreater (1) \textless table\textgreater (1) \newline
The prompt (error information) would be: 
The sub-task: "[PUTBACK] \textless plate\textgreater (1) \textless table\textgreater (1)" caused an error: Script is not executable, since \textless character\textgreater (1) is not close to \textless table\textgreater (1) when executing "[PUTBACK] \textless plate\textgreater (1) \textless table\textgreater (1) [1]" 
Among the following actions, which action would you take. \newline
A. [Find] \textless table\textgreater (1) \newline
B. [Find] \textless plate\textgreater (1) \newline
A corrective choice of sub-task would be (You just need to provide the mark before the option you want to choose): A 
}

\textbf{\textit{(Observation)}}
\textcolor[rgb]{0,0.392,0}{
Currently, you are standing in the bedroom, and holding nothing in your right hand and nothing in your left hand.
 pillow is clean. napkin is clean. pillow is dirty. bed is clean. mat is dirty. pillow is close to drawing. tablelamp is close to bed. mat is facing drawing. pillow is inside bedroom. mat is close to table. bed is close to drawing. table is close to mat. pillow is on floor. floor is close to bed. tablelamp is close to pillow. mat is close to curtain. bed is facing computer. mat is close to floor. pillow is facing drawing. curtain is close to mat. bed is close to tablelamp. wall is close to mat. napkin is inside bedroom. window is close to mat. pillow is close to tablelamp. bed is close to floor. pillow is close to wall. mat is close to filing\_cabinet. drawing is close to bed. pillow is close to floor. bed is close to wall. filing\_cabinet is close to mat. bed is close to nightstand. mat is inside bedroom. pillow is close to pillow. pillow is close to nightstand. mat is close to wall. wall is close to pillow. nightstand is close to pillow. nightstand is close to bed. drawing is close to pillow. floor is close to pillow. bed is inside bedroom. floor is close to mat. wall is close to bed. mat is close to window. pillow,napkin,pillow,mat,bed,pillow,pillow is inside bedroom.}

Your task is: Take nap. 

\textbf{\textit{(History)}}\newline
\textcolor[rgb]{0,0,0.545}{
Your previously executed sub-tasks are:\newline
[\textbf{Walk}] \textless bedroom\textgreater (1)
}

Among the following sub-tasks, which one would you take.\newline
A. [FIND] \textless bed\textgreater (1)\newline
B. [WALK] \textless couch\textgreater (1) \newline
C. [WALK] \textless bed\textgreater (1) \newline
D. [FIND] \textless couch\textgreater (1)\newline
E. [FIND] \textless pillow\textgreater (1)\newline
The best choice of sub-task is: 



\section{Visualized Action Tree}
\label{appendix:viz-decision-tree}
We visualized the action trees for the tasks listed in Table~\ref{tab:action-trees} for interested readers.

\begin{table}[ht]
\centering
\begin{tabular}{lcc}
\toprule
\textbf{Task} & \textbf{Action Tree}\\
\midrule
\textit{Take nap} & See Figure~\ref{fig:decision-tree-take-nap}\\
\textit{Put on your shoes} & See Figure~\ref{fig:decision-tree-put-on-your-shoes}\\
\textit{Pick up spare change on dresser} & See Figure~\ref{fig:decision-tree-pick-up-spare-change-on-dresser}\\
\textit{Clean toilet} & See Figure~\ref{fig:decision-tree-clean-toilet}\\
\textit{Put alarm clock in bedroom} & See Figure~\ref{fig:decision-tree-put-alarm-clock-in-bedroom}\\
\bottomrule
\end{tabular}
\caption{Action trees for various tasks.}
\label{tab:action-trees}
\end{table}

\begin{figure}[ht]
    \centering
    \includegraphics[width=1\textwidth]{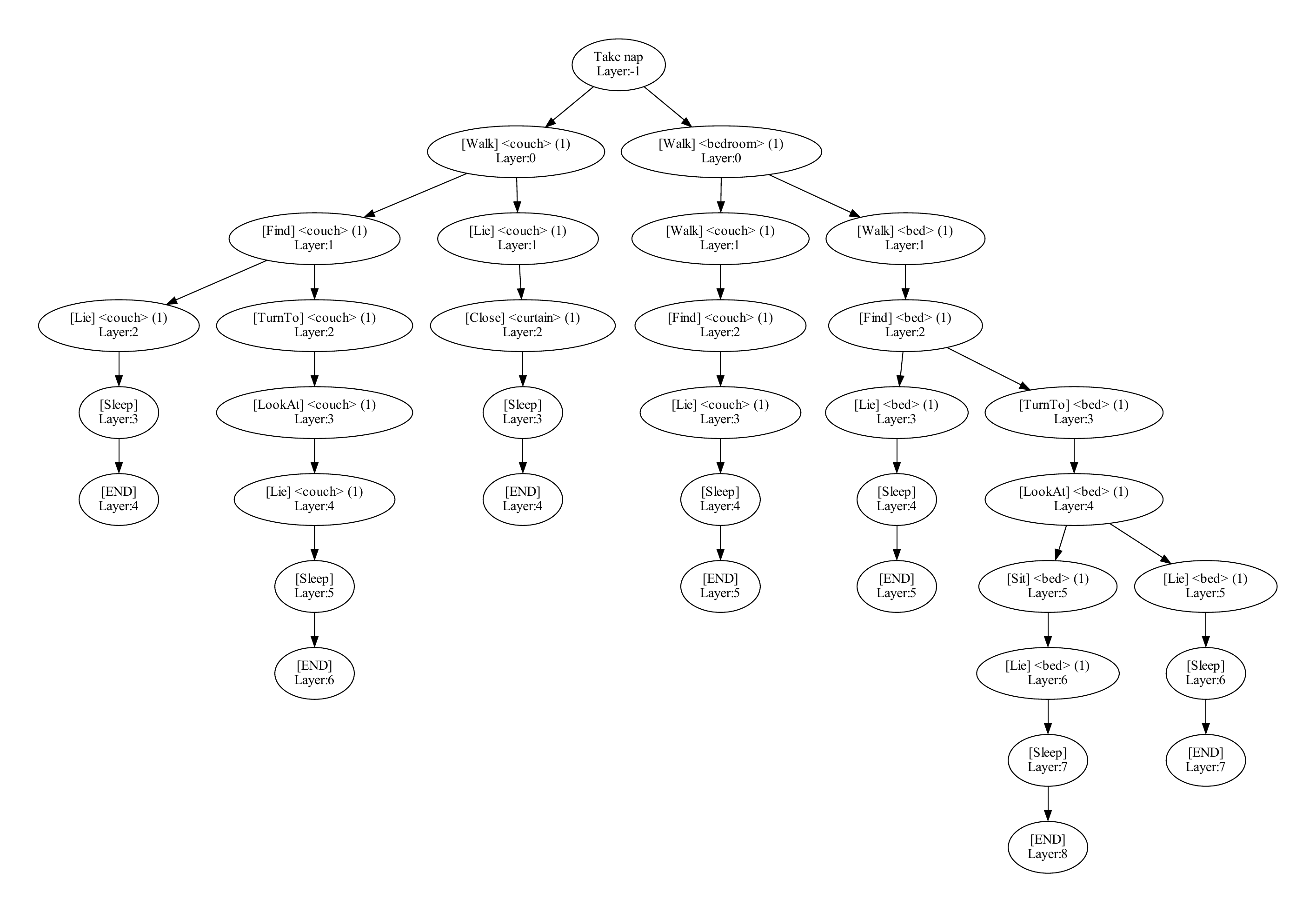}
\caption{Visualization of the action tree for \textit{Take nap}.}
\label{fig:decision-tree-take-nap}
\end{figure}

\begin{figure}[ht]
    \centering
    \includegraphics[width=1\textwidth]{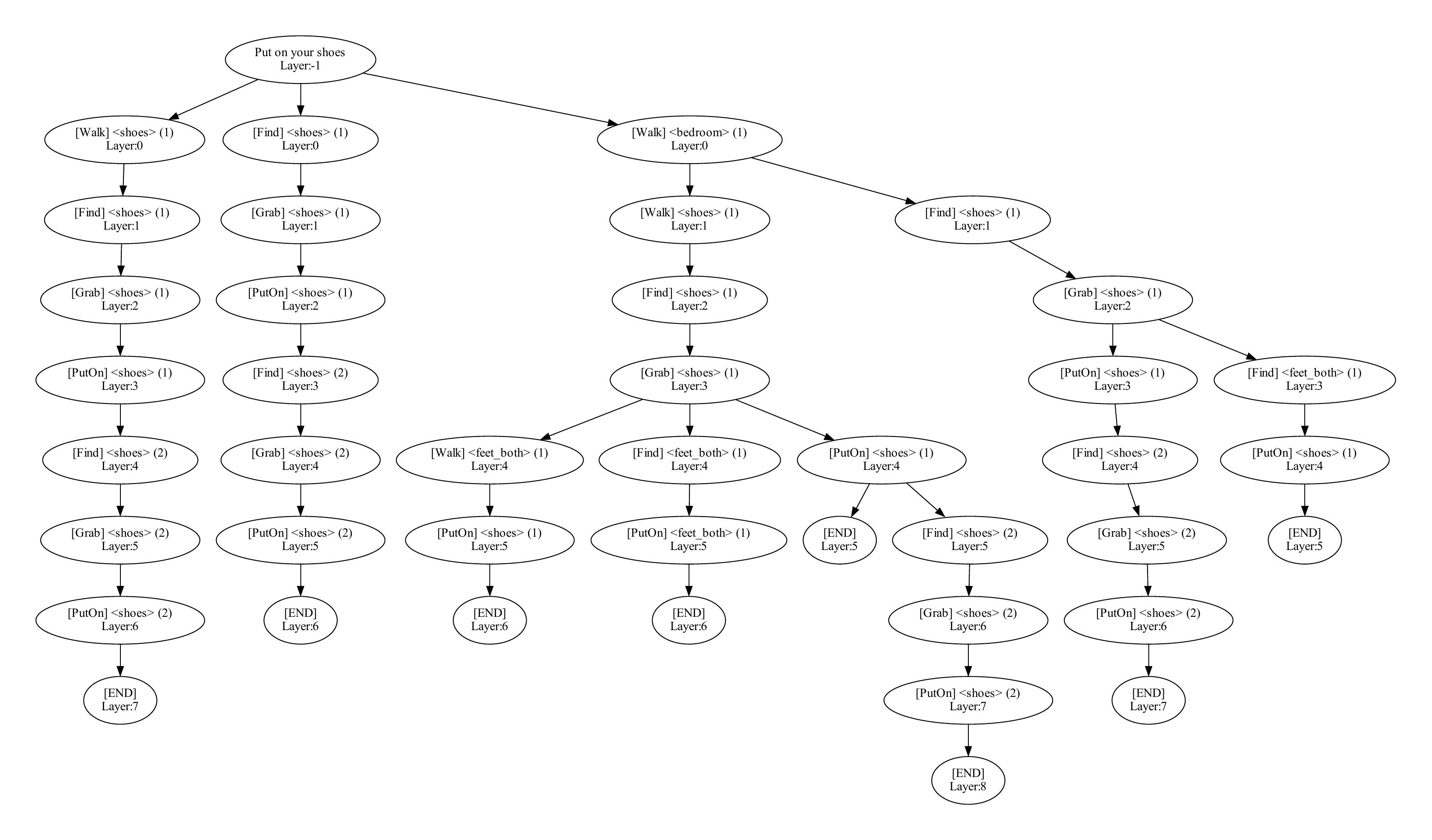}
\caption{Visualization of the action tree for \textit{Put on your shoes}.}
\label{fig:decision-tree-put-on-your-shoes}
\end{figure}

\begin{figure}[ht]
    \centering
    \includegraphics[width=1\textwidth]{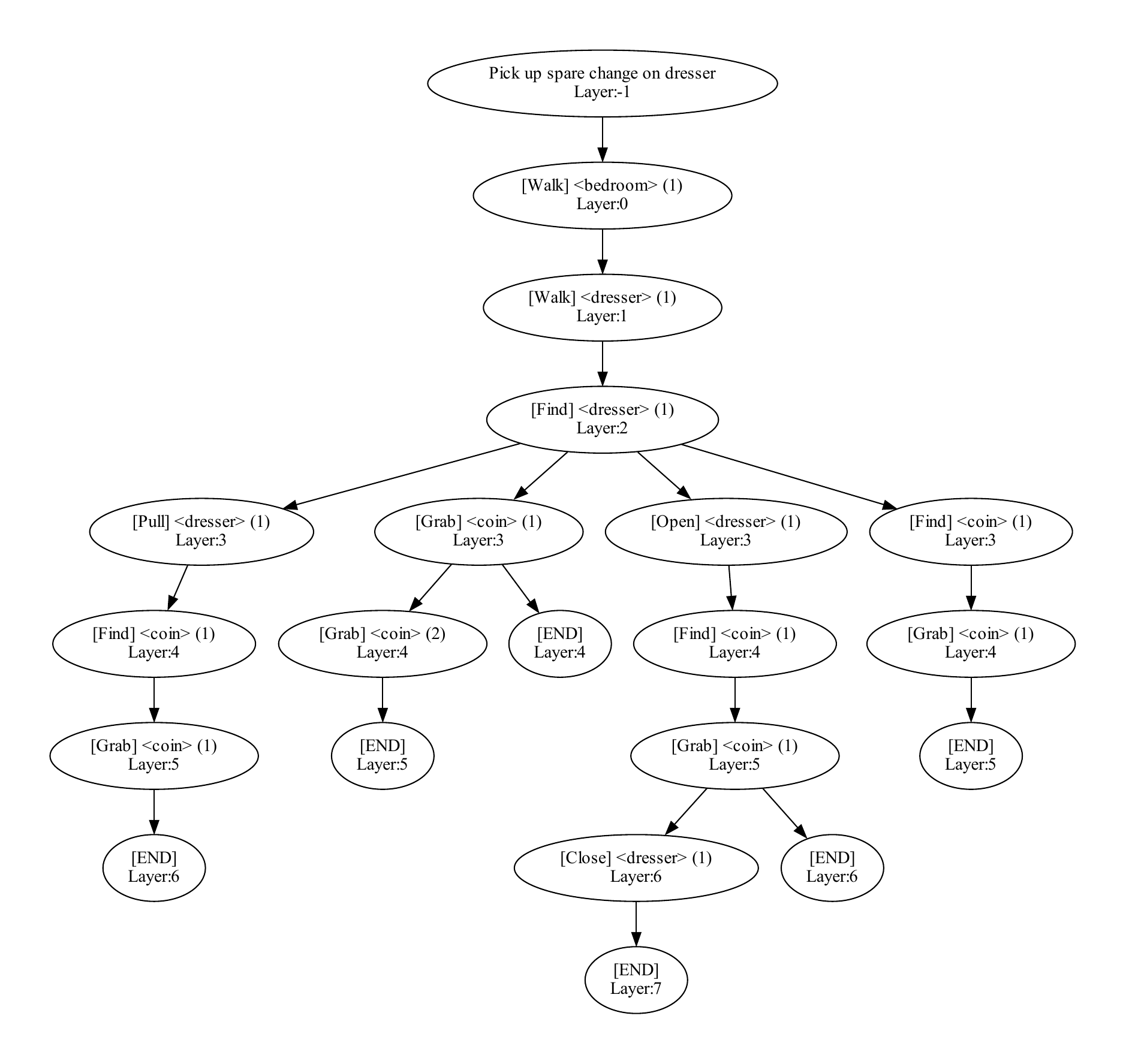}
\caption{Visualization of the action tree for \textit{Pick up spare change on dresser}.}
\label{fig:decision-tree-pick-up-spare-change-on-dresser}
\end{figure}

\begin{figure}[ht]
    \centering
    \includegraphics[width=1\textwidth]{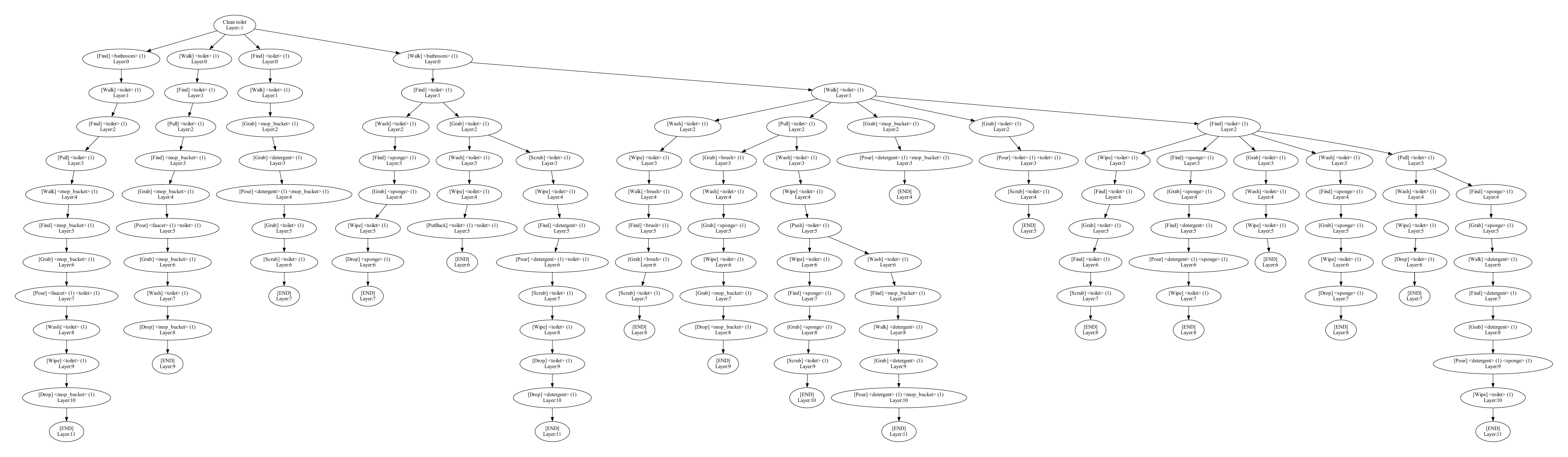}
\caption{Visualization of the action tree for \textit{Clean toilet}.}
\label{fig:decision-tree-clean-toilet}
\end{figure}

\begin{figure}[ht]
    \centering
    \includegraphics[width=1\textwidth]{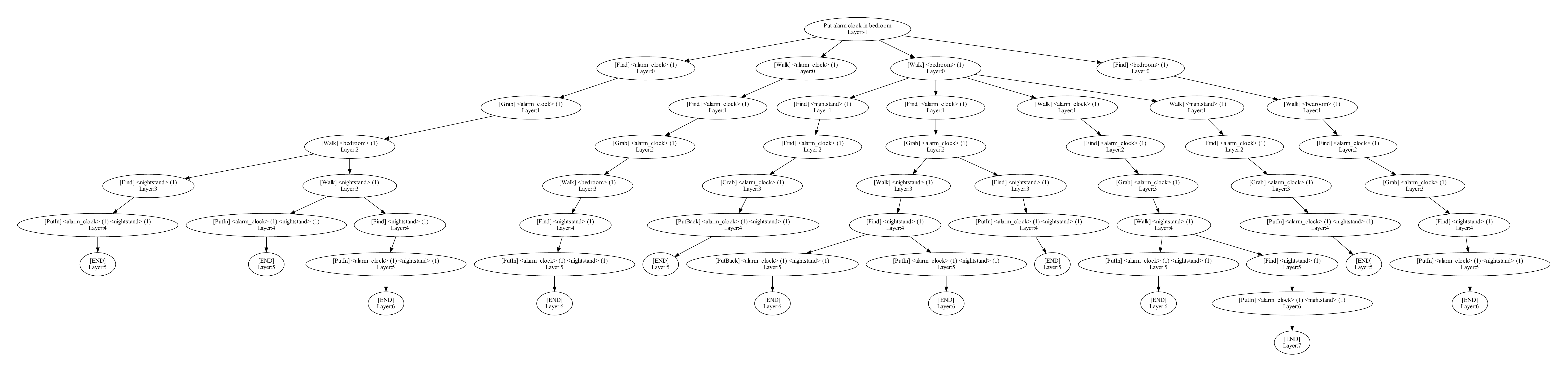}
\caption{Visualization of the action tree for \textit{Put alarm clock in bedroom}.}
\label{fig:decision-tree-put-alarm-clock-in-bedroom}
\end{figure}

\end{document}